\documentclass[dvipsnames,11pt]{article}
\usepackage{charter}
\usepackage{fullpage}
\usepackage{parskip}

\usepackage{amsmath}
\usepackage{amssymb}
\usepackage{booktabs}
\usepackage{color}
\usepackage{colortbl}
\usepackage{enumitem}
\usepackage{epsfig}
\usepackage{etoolbox}
\usepackage{graphicx}
\usepackage[utf8]{inputenc}
\usepackage{listings}
\usepackage{makecell}
\usepackage{multirow}
\usepackage[numbers,sort&compress]{natbib}
\usepackage[olditem,oldenum]{paralist}
\usepackage{tabularx}
\usepackage{wrapfig}
\usepackage{subcaption}
\usepackage[dvipsnames]{xcolor}
\usepackage{xspace}

\usepackage[pagebackref,breaklinks,colorlinks,bookmarks=False]{hyperref}
\usepackage{inconsolata}
\usepackage{xparse}

\renewcommand{\paragraph}[1]{\vspace{2pt} \noindent \textbf{#1}}
\newcommand{\dsquestion}[1]{\emph{#1}}
\newcommand{\dsanswer}[1]{\textcolor{OliveGreen}{#1}}
\newcommand{\qref}[1]{\textcolor{Red}{Q}\ref{#1}}

\renewcommand{\arraystretch}{1.2}
\def\eg{\emph{e.g.,}\xspace} 
 
\hypersetup{
    urlcolor=blue,
    citecolor=ForestGreen,
}

\usepackage[capitalize,noabbrev]{cleveref}

\title{Benchmarking Object Detectors with COCO:\\A New Path Forward}

\author{
    \newcolumntype{Z}{>{\centering\arraybackslash}X}
    \begin{tabularx}{0.8\linewidth}{ZZZ}
        Shweta Singh\textsuperscript{1$\star$} &
        Aayan Yadav\textsuperscript{1$\star$} &
        Jitesh Jain\textsuperscript{2} \\
        Humphrey Shi\textsuperscript{2} &
        Justin Johnson\textsuperscript{3} &
        Karan Desai\textsuperscript{3} \\
    \end{tabularx}
    \\
    \small
    \textsuperscript{1}IIT Roorkee \ \ \ \ \
    \textsuperscript{2}Georgia Tech \ \ \ \ \
    \textsuperscript{3}University of Michigan \\
}
\date{}  

\begin{document}
\maketitle

\begin{abstract}
The Common Objects in Context (COCO) dataset has been instrumental in benchmarking object detectors over the past decade.
Like every dataset, COCO contains subtle errors and imperfections stemming from its annotation procedure.
With the advent of high-performing models, we ask whether these errors of COCO are hindering its utility in reliably benchmarking further progress.
In search for an answer, we inspect thousands of masks from COCO (2017 version) and uncover different types of errors such as
imprecise mask boundaries, non-exhaustively annotated instances, and mislabeled masks.
Due to the prevalence of COCO, we choose to correct these errors to maintain continuity with prior research.
We develop \textbf{COCO-ReM} (\textbf{Re}fined \textbf{M}asks), a cleaner set of annotations with visibly better mask quality than COCO-2017.
We evaluate fifty object detectors and find that models that predict visually sharper masks score higher on COCO-ReM,
affirming that they were being incorrectly penalized due to errors in COCO-2017.
Moreover, our models trained using COCO-ReM converge faster and score higher than their larger variants trained using COCO-2017,
highlighting the importance of data quality in improving object detectors.
With these findings, we advocate using COCO-ReM for future object detection research.
Our dataset is available at \url{https://cocorem.xyz}
\end{abstract}

\renewcommand*{\thefootnote}{$\star$}
\setcounter{footnote}{1}
\footnotetext{Denotes equal contribution. Correspondence to: Karan Desai (\texttt{kdexd@umich.edu})}
\renewcommand*{\thefootnote}{\arabic{footnote}}
\setcounter{footnote}{0}

\section{Introduction}
\label{sec:cocorem_introduction}

Rigorous benchmarking of computer vision models relies on task-specific datasets annotated by humans \citep{imagenet,lin2014coco,cityscapes,ade20k}.
Understandably, human-annotated data is prone to errors and imperfections due to factors such as
ambiguous instructions to annotators, differences in perspective of multiple annotators, \emph{etc.}
Errors in ground-truth training data provide a noisy learning signal for models, resulting in sub-par capabilities.
Even more problematic is the presence of errors in ground-truth evaluation data,
which can cause a misalignment between evaluation metrics and human judgment when comparing model performances.
In this work, we comprehensively study the extent of this issue for the canonical computer vision task of object detection and segmentation.

\begin{figure}[t]
    \centering
    \includegraphics[width=0.99\linewidth]{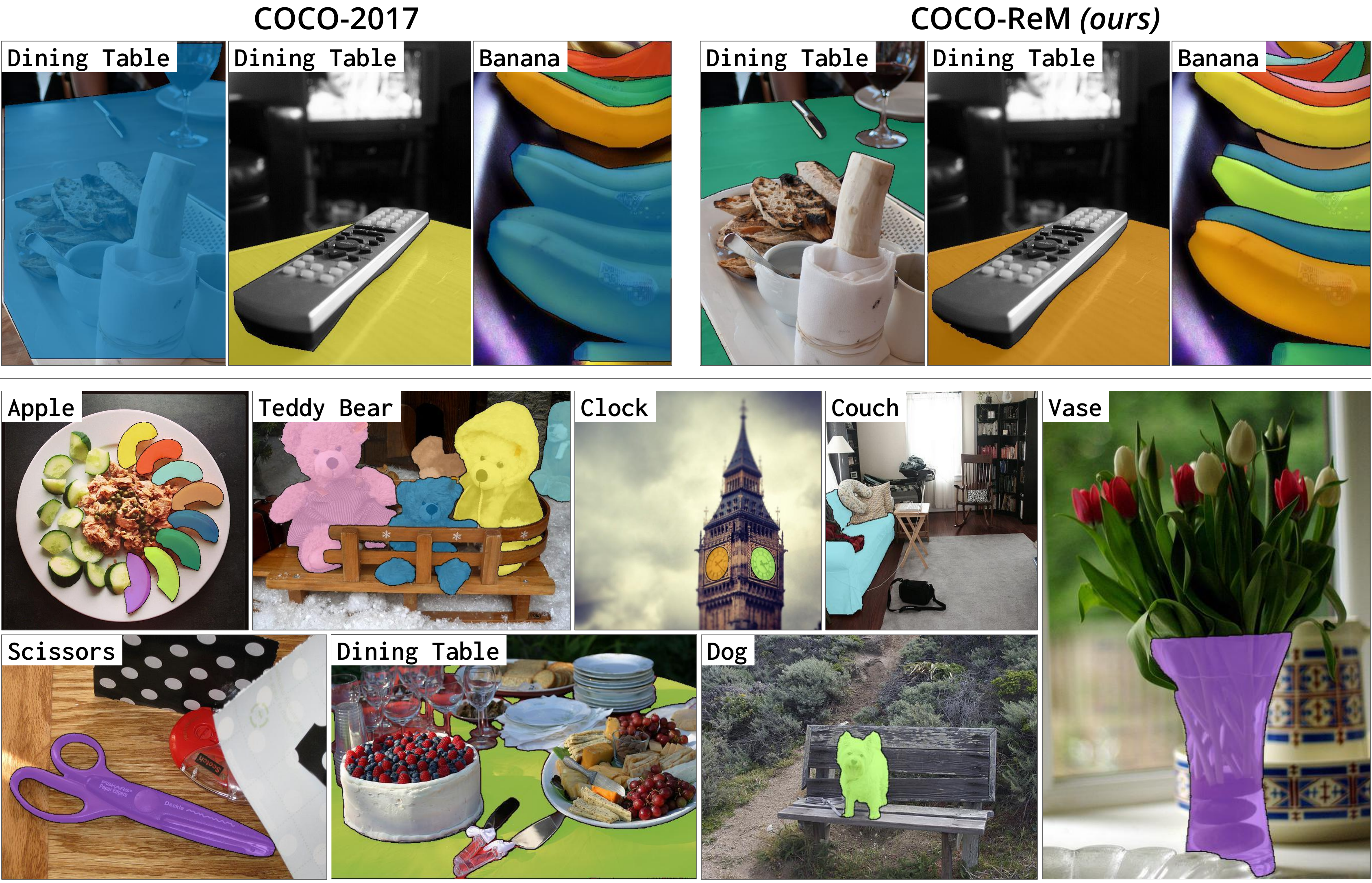}
    \vspace{-5pt}
    \caption{
        \textbf{COCO with Refined Masks (COCO-ReM).}
        \textbf{Top:} COCO-ReM improves upon the quality of COCO-2017 masks by handling occlusions consistently (\emph{dining table}) and
        annotating instances exhaustively (\emph{banana}).
        \textbf{Bottom:}
        Random examples from COCO-ReM showing the mask quality (single category per image, for clarity).
    }
    \label{fig:intro_figure}
\end{figure}

The Common Objects in Context (COCO) dataset~\citep{lin2014coco} has been the standard benchmark for object detection and segmentation since its inception in 2014.
COCO fostered the empirical study of modern object detectors by providing a large set of 123K images annotated with 896K instance masks.
The Mask AP (average precision) of COCO detectors has improved dramatically,
with $\approx$70\% \emph{relative gain} between the pioneering Mask R-CNN~\citep{he2017mask,resnet} from 2017,
and recent models published in 2022--23 such as ViTDet~\citep{cai2018cascade,li2022vitdet}
and other Transformer-based models \citep{vaswani2017attention, detr, dosovitskiy2021vit} like Mask2Former~\citep{cheng2022mask2former} and OneFormer~\citep{jain2022oneformer}.

Despite its popularity, COCO is not without limitations.
COCO masks are known to have \emph{coarse} boundaries~\citep{gupta2019lvis,benenson2019lopenimages}.
We manually inspect thousands of masks in COCO (2017 version) and observe other imperfections.
As shown in \Cref{fig:intro_figure} (top), COCO-2017 masks do not handle holes and occlusions properly (\emph{dining table}),
and sometimes have non-exhaustively annotated instances (\emph{banana}).
We elaborate on these flaws in \Cref{sec:cocorem_revisit}.

From an evaluation perspective, imperfections in \emph{ground-truth} data make the metrics less reliable.
COCO-2017 AP would wrongfully penalize models that predict more precise masks than the imperfect \emph{ground-truth} masks.
Moreover, models trained with such imperfect masks may exploit unwanted biases from the training data,
\eg{} learning to \emph{never} predict masks with holes.
Recently, \citet{kirillov2023sam} observed that human evaluators consistently rated the masks predicted by their Segment Anything Model (SAM)
to be of higher quality than a ViTDet model~\citep{li2022vitdet} trained on COCO, whereas the latter scored higher on COCO AP.

These observations raise an important question -- \emph{can COCO reliably benchmark the future progress in object detection research?}
COCO is the \emph{de-facto} benchmark for object detection -- countless research papers report COCO results every year;
libraries like Torchvision~\citep{torchvision2016} and Detectron2~\citep{wu2019detectron2} readily support object detector development with COCO.
Hence, to maintain continuity with prior research and tap into this rich open-source ecosystem,
we choose to rectify the errors of COCO and reinforce its utility for future research.

\paragraph{Contributions.}
We develop \textbf{COCO-ReM} (\textbf{Re}fined \textbf{M}asks), a cleaner set of high-quality instance annotations for COCO images.
The improved quality over COCO-2017 masks is \emph{visually palpable}, as seen in \Cref{fig:intro_figure}.

Our annotation pipeline (\Cref{sec:cocorem_dataset}) systematically rectifies the imperfections of COCO-2017 masks.
We refine the mask boundaries using SAM~\citep{kirillov2023sam},
and exhaustively annotate instances by importing from the LVIS dataset~\citep{gupta2019lvis} and
an ensemble of LVIS-trained models.
\emph{We (the authors) manually verify the entire validation set to provide a strong quality guarantee for evaluation.}

In \Cref{sec:cocorem_experiments}, we use COCO-ReM for benchmarking object detectors.
We evaluate fifty object detectors and observe that their COCO-ReM AP produces a different model ranking than COCO-2017 AP.
Surprisingly, we observe that \emph{query-based} models (Mask2Former and OneFormer) score much higher on COCO-ReM than \emph{region-based} models (ViTDet).
Query-based models predict visually sharper masks -- AP trends on COCO-ReM accurately reflect this, unlike COCO-2017.
We observe that models trained using COCO-ReM converge faster and perform better than those trained using COCO-2017,
highlighting the importance of mask quality in improving object detectors.

\section{Background: Revisiting COCO Masks}
\label{sec:cocorem_revisit}

The annotation pipeline of a dataset determines the quality of annotations obtained.
To design an effective annotation pipeline for COCO-ReM,
we review the COCO annotation pipeline by \citet{lin2014coco}
and understand its shortcomings.

\paragraph{COCO annotation pipeline.}
\citet{lin2014coco} employed a three-stage pipeline to collect instance mask annotations for COCO images:
\begin{compactenum}[\hspace{1pt}1.]
    \item \textbf{Category labeling:}
    Annotators assigned one or more labels to images depending on the visible content, choosing from a fixed set of categories.
    \item \textbf{Instance spotting:}
    Annotators locate up to ten instances per label (from the previous stage) in the corresponding image, by providing clicks (points).
    \item \textbf{Instance segmentation:}
    Finally, annotators clicked around the boundaries of each located instance (from the previous stage), rendering a \emph{polygon} mask.
\end{compactenum}
\citet{lin2014coco} also took numerous measures to improve annotation quality.
The first two stages -- \emph{category labeling} and \emph{instance spotting} -- were performed by eight annotators per image to maximize recall.
The \emph{instance segmentation} stage required annotators to complete a training task before annotating actual instances in the dataset.
Next, a different set of crowd workers verified the correctness and quality of annotated masks.
Finally, the annotators used a \emph{paintbrush} tool to mark remaining objects as \emph{crowd regions}.
These extensive measures avoided many scenarios that could have led to low-quality masks.

\begin{figure}[t]
    \includegraphics[width=\linewidth]{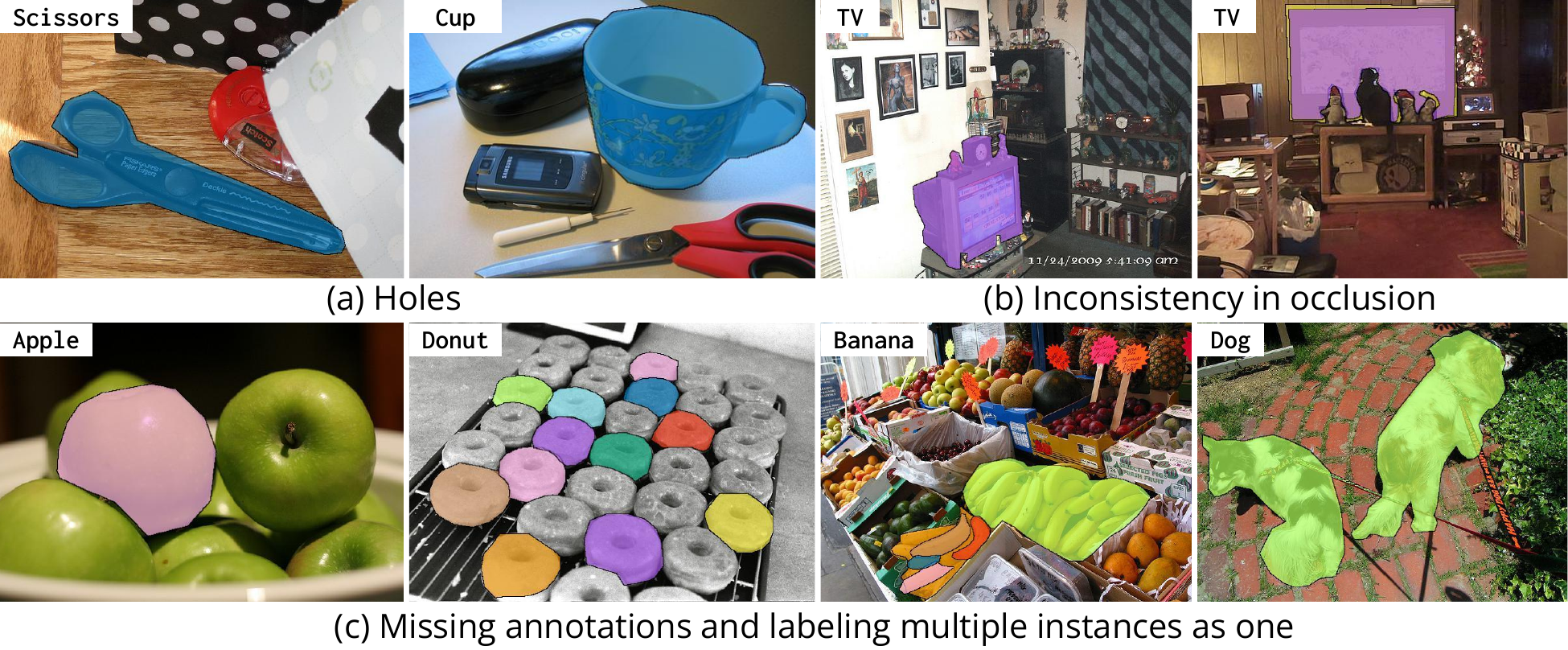}
    \vspace{-20pt}
    \caption{
        \textbf{Imperfections in COCO-2017 masks.}
        All masks in general have coarse boundaries.
        Moreover, they lack holes when required (\emph{scissors})
        and do not handle occlusions consistently (\emph{TV} with or without figurines).
        Annotations are sometimes non-exhaustive, either masks are missing (\emph{apples} and \emph{donuts}),
        or multiple instances are grouped into a single mask (\emph{banana} and \emph{dog}).
    }
    \label{fig:coco_inconsistencies}
\end{figure}

\paragraph{Imperfections in COCO-2017 masks.}
To further understand the flaws in COCO annotations,
we visualized thousands of COCO-2017 masks and inspected their quality and label correctness.
Through this exercise, we observed recurring types of errors in annotations,
as shown in \Cref{fig:coco_inconsistencies}.

\begin{compactenum}[\hspace{1pt}1.]
    \item \textbf{COCO-2017 masks have coarse boundaries.}
    This is a natural consequence of collecting masks as \emph{polygon contours},
    which only coarsely approximate the pixel-precise mask boundaries through piece-wise straight edges connecting points (clicks) supplied by annotators.
    This shortcoming is well-known and identified by several follow-up works~\citep{gupta2019lvis,benenson2019lopenimages,kirillov2023sam}.
    \item \textbf{COCO-2017 masks lack holes when required.}
    The polygon-drawing interface of COCO did not allow annotators to \emph{erase} the interior regions of polygons.
    Consequently, certain objects that should naturally contain holes, lack them.
    \Cref{fig:coco_inconsistencies}\textcolor{red}{a} shows two examples -- the handles of \emph{scissors} and \emph{cup}.
    While there should be holes in the masks, COCO-2017 does not include them.
    \item \textbf{COCO-2017 masks do not handle occlusions consistently.}
    Masks of occluded objects are either drawn \emph{over} the other occluding objects (\emph{amodal} masks) or \emph{around} them (\emph{modal} masks).
    For example, observe the \emph{TV} masks in \Cref{fig:coco_inconsistencies}\textcolor{red}{b}.
    While both, \emph{amodal} and \emph{modal} masks are valid, either one should be consistent in the dataset according to the task definition.
    We believe this inconsistency is a consequence of inadequate task instructions to annotators as well as the lack of holes in masks.
    \item \textbf{COCO-2017 masks are often non-exhaustive.}
    This issue occurs in two modes, as shown in \Cref{fig:coco_inconsistencies}\textcolor{red}{c} --
    either few instance masks of a category are missing\footnote{These images have all their missing masks annotated as \emph{non-crowd} regions.}
    (\emph{apple}, \emph{donut}), or multiple instances are grouped in a single mask (\emph{banana}, \emph{dog}).
    This is a recurring issue throughout COCO --
    we believe that certain annotators may not have deemed masks with grouped instances to have a \emph{wrong object contour}, due to oversight.
    \item \textbf{COCO-2017 contains few near-duplicate masks.}
    We found 410 pairs of masks that overlap with intersection-over-union (IoU) greater than 0.8 in the COCO-2017 validation set ($\approx$2.3\% of all instances).
    These pairs often segment the same underlying object with different labels.
    Examples include armchairs (\emph{chair} and \emph{couch}), calves (\emph{cow} and \emph{sheep}), and duffel bags (\emph{suitcase} and \emph{backpack}).
    We attribute this issue to the lack of a cross-category de-duplication step in the COCO annotation pipeline.
\end{compactenum}

\noindent All these imperfections hurt the reliability of the benchmarking process.
They provide noisy supervision during training and wrongfully penalize correct model predictions during evaluation.
Having characterized these imperfections, we make targeted design choices in our COCO-ReM annotation pipeline to rectify them.

\section{COCO-ReM: COCO with Refined Masks}
\label{sec:cocorem_dataset}

In this section, we outline our annotation process for \textbf{COCO-ReM}.
Our goal is to rectify the imperfections in COCO-2017 annotations as discussed in \Cref{sec:cocorem_revisit} to maintain continuity with prior research.
We hope that higher quality \emph{ground-truth} masks would improve the reliability of the benchmarking process.


\subsection{Annotation Pipeline}
\label{subsec:cocorem_pipeline}

\begin{figure}[t]
    \includegraphics[width=\linewidth]{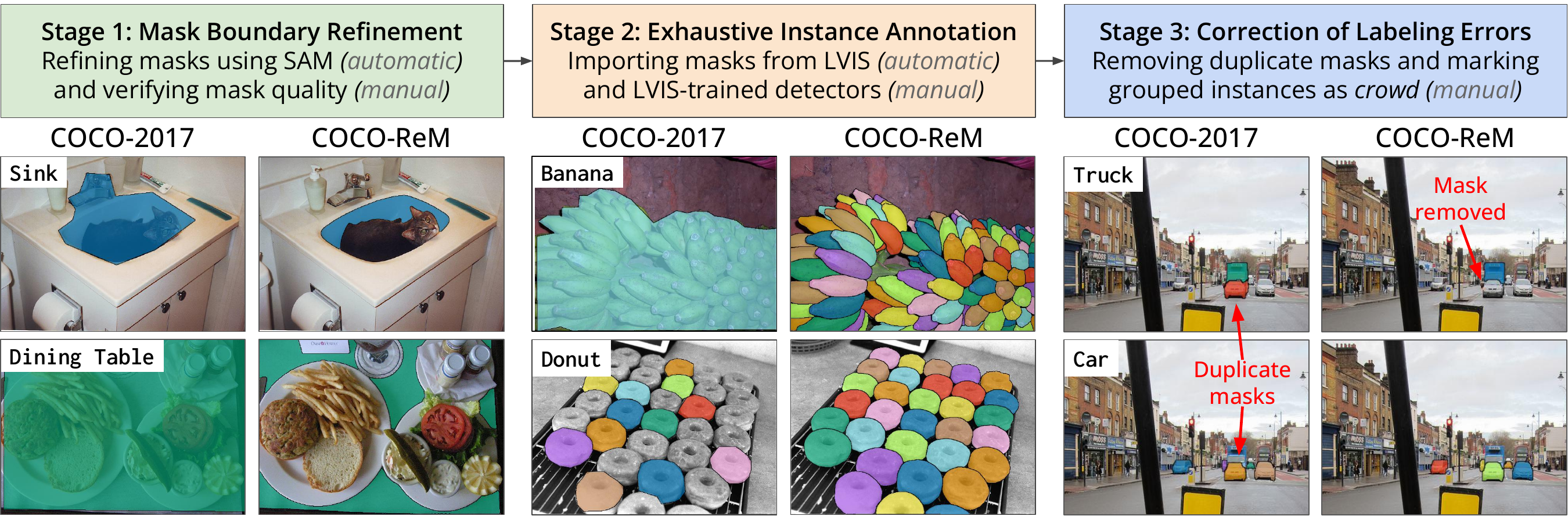}
    \vspace{-15pt}
    \caption{
        \textbf{COCO-ReM annotation pipeline.}
        We develop a \emph{semi-automatic} annotation pipeline comprising three stages to rectify the imperfections in COCO-2017 masks.
    }
    \label{fig:cocorem_ann_pipeline}
\end{figure}

We develop a \emph{semi-automatic} pipeline --
first, we refine the masks using off-the-shelf models and LVIS dataset~\cite{gupta2019lvis},
followed by manual verification of the updated masks.
We refine both, the training and validation set of COCO-2017, comprising $\approx$860K and $\approx$36K instances respectively.
We thoroughly verify the validation set to ensure \emph{consistency} and \emph{pixel-precise mask quality} for model evaluation.
As the training set is much larger, it is only processed through the automatic steps of our pipeline.
Our pipeline comprises three stages (\Cref{fig:cocorem_ann_pipeline}).


\paragraph{Stage 1. Mask boundary refinement.}
In this stage, we aim to rectify the coarse boundary quality of COCO-2017 masks.
A straightforward solution to achieve this is to re-annotate all masks
through a web interface with \emph{paintbrush} and \emph{eraser} tools instead of drawing polygon contours.
Moreover, this solution naturally allows including holes in masks where necessary.
While accurate, this solution would sharply increase the mask annotation time and cost.

Along a different trend, prior works have explored \emph{model-assisted interactive segmentation} as a feasible solution for collecting mask annotations with smooth boundaries~\citep{benenson2019lopenimages,gupta2019lvis}.
The key idea is to let an object detector predict a coarse initial mask
and have the annotators refine it by prompting corrective mouse clicks.
Inspired by the accuracy and practical feasibility of this approach,
we use the Segment Anything Model (SAM~\citep{kirillov2023sam}) to refine the COCO-2017 masks.

\begin{compactitem}[\hspace{1pt}--]
    \item \textbf{Refining masks using SAM.}
    SAM is a promptable segmentation model that predicts high-quality masks conditioned on user-provided point and box prompts.
    This promptable functionality makes SAM suitable for our use case.
    We prompted SAM using various combinations of prompts sampled using COCO-2017 masks and obtained its predictions as the refined masks.
    Refer to the appendix for details about this prompting methodology.
    \item \textbf{Manual verification of mask quality.}
    We visualized and manually inspected every mask created in the previous process.
    Through this inspection, we identified nearly 900 masks that have low quality or do not accurately cover the underlying object.
    These masks represent only 2.4\% of 36K instances in the COCO-2017 validation set,
    greatly reducing our manual effort and showing the efficacy of our automated process.
    To correct these masks, we deployed a web browser demo of SAM
    and provided point prompts iteratively until obtaining a precise and accurate mask.
\end{compactitem}

\noindent Through this stage, we obtain masks with more precise boundaries and better handling of occlusions, as illustrated in \Cref{fig:cocorem_ann_pipeline} \textcolor{red}{(left)}.

\paragraph{Stage 2. Exhaustive instance annotation.}
This stage is aimed at alleviating the issue of non-exhaustively annotated instances in COCO-2017 (\Cref{fig:coco_inconsistencies}\textcolor{red}{c}).
We note that manually re-annotating every image is the most reliable solution to ensure exhaustive instance coverage.
However, the manual effort would not only require extensive instructions and verification but would also be highly cost-ineffective and may lead to redundant annotations.
To avoid redundancy and annotation costs, we automatically import instances from the LVIS~\citep{gupta2019lvis} dataset as a strong starting point.

\begin{compactitem}[\hspace{1pt}--]
    \item \textbf{Importing instances from LVIS.}
    The LVIS dataset comprises 1.2M high-quality instance annotations across 1203 categories for images in COCO, including all 80 COCO categories.
    \citet{gupta2019lvis} implemented specific instructions and verification steps to firmly ensure exhaustive annotations.
    Every image is labeled with a set of \emph{positive} LVIS categories whose instances are exhaustively labeled,
    and a set of \emph{negative} LVIS categories that are verifiably absent.
    The presence (or absence) of the remaining LVIS categories is \emph{unverified} for the image.

    For the 80 COCO categories, LVIS contains $\approx$20K instances for images in the COCO-2017 validation set.
    To import these instances, we checked whether LVIS contained more instances of an \emph{(image, category)} pair than COCO-2017.
    If so, we replaced all COCO-2017 instances of this pair with those from LVIS.
    Nearly 9\% of \emph{(image, category)} pairs in LVIS are flagged to have non-exhaustive instance annotations~\cite{gupta2019lvis} -- we ignored these while importing.
    \item \textbf{Discovering instances using LVIS-trained models.}
    Instances of a given \emph{(image, category)} pair of COCO-2017 may be missing in LVIS
    if the category is \emph{unverified} for the corresponding image.
    Instances of this pair may be non-exhaustive in COCO-2017, and hence remain unresolved after the previous automatic step.
    To resolve this, we use publicly available object detectors trained using LVIS.
    On the validation set, we ensemble mask predictions using the official checkpoints of three best-performing ViTDet models~\cite{li2022vitdet}.
    As model predictions are not as accurate as human-annotated data,
    we manually inspect $\approx$2500 masks predicted with high confidence and select 1056 masks.
    We refine these masks (Stage 1) and add them to COCO-ReM.
\end{compactitem}

\noindent This stage augments COCO-ReM with additional instances from external sources
and includes more exhaustive annotations, see example in \Cref{fig:cocorem_ann_pipeline} \textcolor{red}{(middle)}.


\paragraph{Stage 3. Correction of labeling errors.}
This is the final stage of our pipeline, performed manually, only for the validation set.
Recall from \Cref{sec:cocorem_revisit} that we found 410 near-duplicate mask pairs in the COCO-2017 validation set.
We manually inspected these and retained one mask per pair with the most accurate label,
such as \emph{chair} for armchairs and \emph{car} for SUVs (\Cref{fig:cocorem_ann_pipeline} \textcolor{red}{(right)}).
We also observed that $\approx$100 grouped instances were not resolved through Stage 2.
We annotated these masks as \emph{crowd} to accurately reflect their characteristics.


\subsection{Mask Characteristics}
\label{subsec:cocorem_mask_chars}

\paragraph{Instance statistics.}
Our annotation pipeline imports instances from external sources -- LVIS dataset and LVIS-trained models -- to annotate instances exhaustively.
As a result, every instance in COCO-ReM can be traced back to either one of these sources or COCO-2017.
As shown in \Cref{tab:cocorem_instance_statistics},
our COCO-ReM validation set extends COCO-2017 by including
$\approx$6K masks from LVIS~\citep{gupta2019lvis} and
$\approx$1K masks from LVIS-trained models~\citep{cai2018cascade,li2022vitdet}.
Since these masks replace a subset of COCO-2017 masks (see Stage 2 description),
we retain $\approx$33.5K masks from COCO-2017 (out of 36K masks).
Overall, COCO-ReM validation set has $\approx$4K more instances than COCO-2017.

\begin{table}[t]
    \centering
    \setlength{\tabcolsep}{6pt}
    \begin{tabularx}{\linewidth}{XX cccc}
        \toprule
        ~ & ~ & \multicolumn{3}{c}{Instance Source} & ~ \\
        \cmidrule{3-5}
        Dataset & Split & COCO-2017 & LVIS & LVIS models & Total \\
        \midrule
        COCO-2017 & \texttt{val} & 36,781 & - & - & 36,781 \\
        COCO-ReM  & \texttt{val} & 33,498 & 6,135 & 1056 & 40,689 \\
        \midrule
        COCO-2017 & \texttt{train} & 860,001 & - & - & 860,001 \\
        COCO-ReM  & \texttt{train} & 738,353 & 354,674 & - & 1,093,027 \\
        \bottomrule
    \end{tabularx}
    \vspace{1pt}
    \caption{
        \textbf{Instance statistics in COCO-ReM.}
        We list the exact number of annotations obtained from different sources for both \texttt{train} and \texttt{val} splits of COCO-ReM.
        All masks undergo refinement through SAM before being included in COCO-ReM.
    }
    \label{tab:cocorem_instance_statistics}
\end{table}

\begin{figure}[t]
    \centering
    \includegraphics[width=0.7\linewidth]{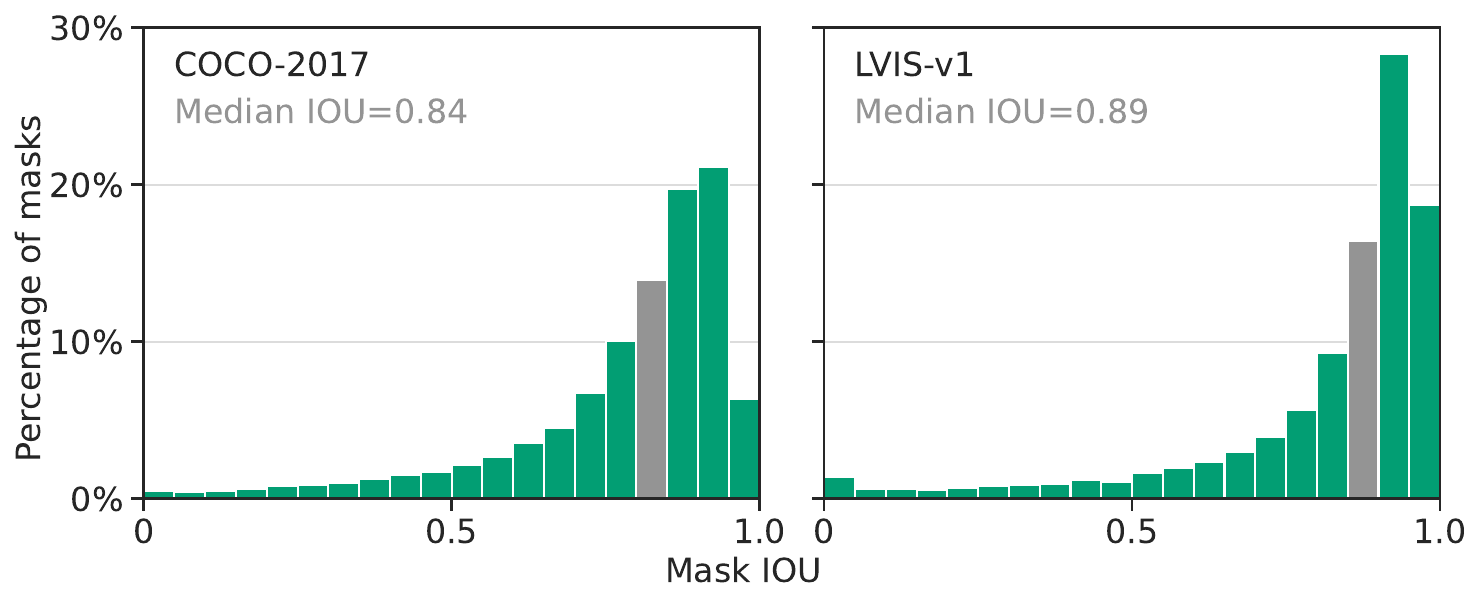}
    \includegraphics[width=0.285\linewidth]{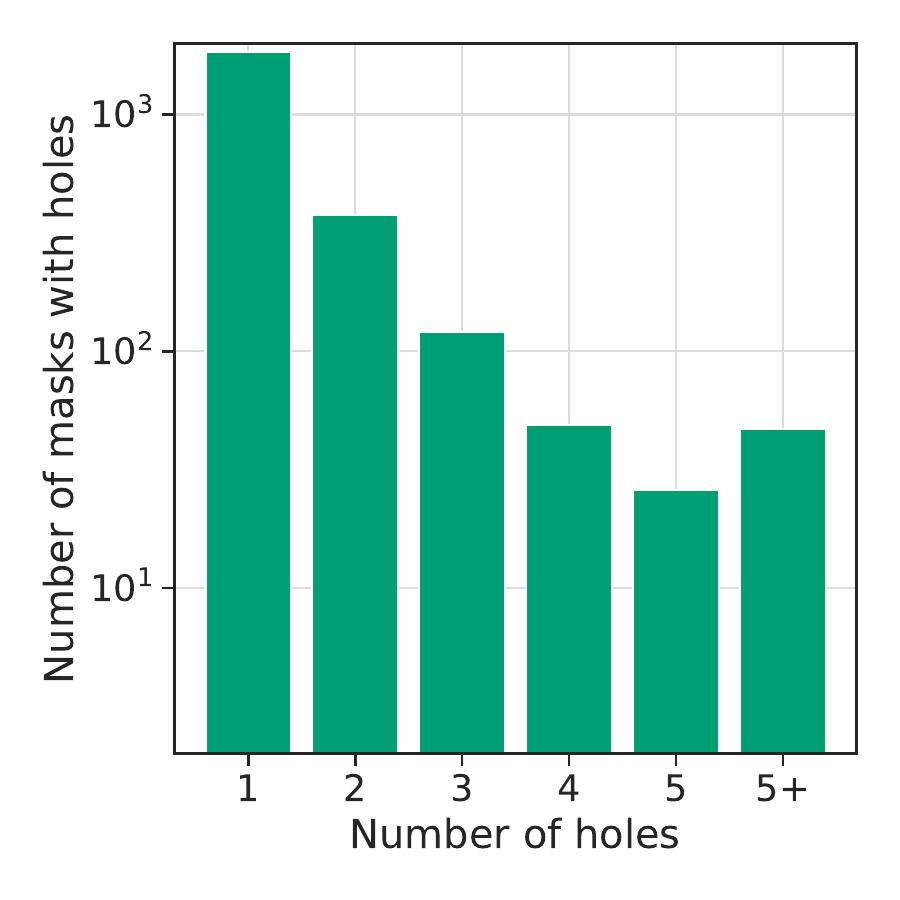}
    \vspace{-15pt}
    \caption{
        \textbf{Left: Mask IoU distribution.}
        Distribution of IoU between masks in COCO-ReM validation set and their source dataset.
        \textbf{Right: Masks with holes.}
        Nearly 2000 masks in COCO-ReM validation set have holes, that were missing in COCO-2017.
    }
    \label{fig:cocorem_mask_iou_holes}
\end{figure}


On the other hand, the training set of COCO-ReM comprises a total of 1.09M high-quality masks,
as compared to 860K masks in the COCO-2017 training set.
The difference in instance counts between COCO-ReM and COCO-2017 sharply highlights the extent of non-exhaustively annotated instances in COCO-2017.

\paragraph{Mask IoU.}
\Cref{fig:cocorem_mask_iou_holes} \textcolor{Red}{(left)} shows the distribution of IoU between the (refined) masks of COCO-ReM (validation set)
and their corresponding source masks in COCO-2017 and LVIS datasets.
The median IoU is slightly less than one, indicating that most masks have a proper coarse shape, but lack precise boundaries.
This median estimate matches the findings of \citet{gupta2019lvis},
albeit observed from a smaller sample of 100 masks collected from human annotators.

\paragraph{Holes in masks.}
\Cref{fig:cocorem_mask_iou_holes} \textcolor{Red}{(right)} shows that $\approx$2000 masks in COCO-ReM validation set have at least one hole.
By design, COCO-2017 lacks these holes.
See the appendix for qualitative examples and additional analysis per category.

\section{Experiments}
\label{sec:cocorem_experiments}

In our experiments, we aim to demonstrate the utility of COCO-ReM for reliable benchmarking of object detection research.
Recall \Cref{sec:cocorem_introduction}, we hypothesize that the imperfections in COCO-2017 may cause two issues:
(1) During evaluation, models may be undesirably penalized despite predicting correct masks,
yielding counter-intuitive modeling observations, and
(2) During training, models receive noisy supervision and/or may learn unwanted biases
(\eg{} never predict masks with holes), resulting in sub-par capabilities.
We claim that high-quality masks of COCO-ReM can alleviate these issues.
To this end, we use COCO-ReM for evaluating and training object detectors
and compare the resulting trends with those obtained using COCO-2017.


\subsection{Evaluation using COCO-ReM}
\label{subsec:cocorem_evaluation}

In this section, we evaluate \emph{fifty} publicly available COCO object detectors using the COCO-ReM validation set.
We cover a wide range of models from existing literature.
All models can be categorized as either \emph{region-based} or \emph{query-based}:
\begin{compactitem}[\hspace{1pt}--]
    \item \textbf{Region-based models}, following Mask R-CNN~\citep{he2017mask}.
    These models first predict labeled boxes, followed by binary masks inside those boxes.
    We also include \emph{cascade} models, following Cascade R-CNN~\citep{cai2018cascade}.
    These models extend the region-based design by iteratively refining their predicted bounding boxes before making the final mask prediction.
    \item \textbf{Query-based models}, inspired by the Detection Transformer design~\citep{detr}.
    Instead of predicting masks within image regions, these models start with a set of image-independent query vectors,
    then decode labeled masks through a Transformer~\citep{vaswani2017attention} by iteratively attending to image embeddings.
    Among query-based models, we consider Mask2Former~\citep{cheng2022mask2former} and OneFormer~\citep{jain2022oneformer}.
\end{compactitem}

\begin{figure}[t]
    \includegraphics[width=\linewidth]{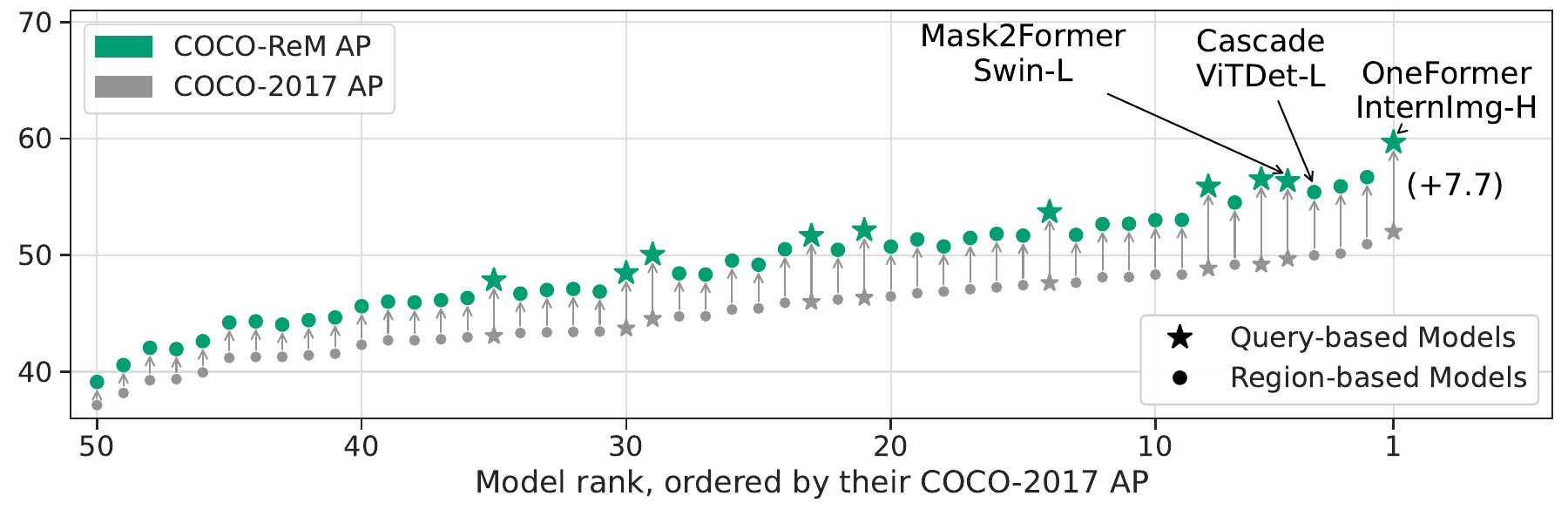}
    \vspace{-20pt}
    \caption{
        \textbf{Evaluating fifty object detectors using COCO-ReM.}
        All models score higher on COCO-ReM than COCO-2017.
        Query-based models ($\bigstar$) score much higher than region-based models on COCO-ReM,
        yielding opposite trends than COCO-2017.
    }
    \label{fig:cocorem_evaluations}
    \vspace{-5pt}
\end{figure}

\noindent We ensure diverse image backbones in our model collection.
The architecture and capacity of the backbone plays a major role in the empirical performance of the object detector.
To this end, we include convolutional backbones (ResNet~\citep{resnet}, RegNet~\citep{Radosavovic2020}, ConvNeXt~\citep{convnext}, InternImage~\citep{wang2022internimage}),
hierarchical vision transformers (Swin~\citep{liu2021swin}, MViTv2~\citep{li2021mvitv2}, DiNAT~\citep{dinat}),
and plain vision transformers~\citep{dosovitskiy2021vit}.
We include backbones of different sizes for a single model,
\eg{} for Cascade ViTDet~\citep{li2022vitdet}, we include ViT-B (141M parameters), ViT-L (361M parameters), ViT-H (692M parameters).

We believe that our model collection is sufficiently diverse to draw meaningful conclusions from comparing trends between COCO-ReM and COCO-2017

\paragraph{Testing protocol.}
We compute the mask Average Precision (AP) of all models using the official COCO evaluation API in Detectron2~\citep{wu2019detectron2}.
For simplicity, we adopt a common testing protocol for all models -- we resize the longest edge of input images to 1024 pixels (without changing the aspect ratio),
and limit the number of detections per image to 100.


\paragraph{Result 1. All models score higher on COCO-ReM.}
In \cref{fig:cocorem_evaluations}, we plot the mask AP trends for all fifty object detectors sorted by their COCO-2017 AP.
All fifty models achieve higher AP on COCO-ReM despite being trained using only COCO-2017 masks.
This result validates our hypothesis that the imperfections of COCO-2017 may have wrongfully penalized all these models.

The difference between COCO-ReM AP and COCO-2017 AP is larger with recent high-performing models --
the current state-of-the-art model (OneFormer InternImage-H) scores \textbf{+7.7} points higher AP on COCO-ReM than COCO-2017.
This increasing gap in AP suggests that the imperfections in ground-truth have worsened the issue of over-penalization
and made COCO-2017 AP less adept at fine-grained assessment and ranking of recent models.

\begin{figure}[t]
    \includegraphics[width=\linewidth]{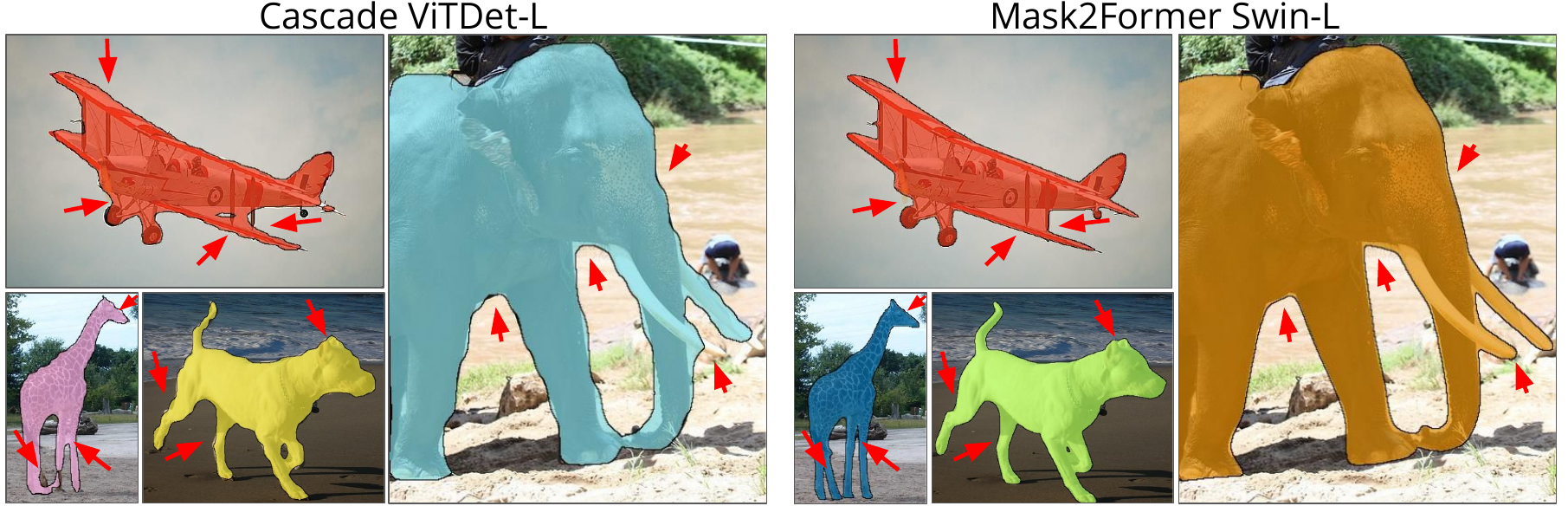}
    \caption{
        \textbf{Mask predictions from models with opposite rank order on COCO-ReM and COCO-2017.}
        Qualitative examples of ViTDet (\emph{region-based} model) and Mask2Former (\emph{query-based} model)
        show that the latter predicts visually sharper masks.
        COCO-ReM AP correctly ranks Mask2Former higher, unlike COCO-2017 AP.
    }
    \label{fig:cocorem_vitdet_mask2former_masks}
\end{figure}

\paragraph{Result 2. Query-based detectors score higher than region-based detectors on COCO-ReM.}
Interestingly, we observe that the trend in AP values calculated using COCO-ReM does not align perfectly with COCO-2017.
We observe frequent \emph{flips} in ranks of several model pairs on COCO-ReM.
While the model rankings do not transfer between benchmarks perfectly,
we observe a consistent pattern in model pairs with flipped rankings.
Notice the star markers ($\bigstar$) in \Cref{fig:cocorem_evaluations} --
the flips are attributed to significantly higher scores of query-based models (Mask2Former~\citep{cheng2022mask2former} and OneFormer~\citep{jain2022oneformer})
than region-based models on COCO-ReM.

To further understand the cause of this performance trend,
we observe the predicted masks of two models having opposite rank order according to COCO-ReM and COCO-2017 AP:
Cascade ViTDet-L~\citep{li2022vitdet} and Mask2Former Swin-L~\citep{cheng2022mask2former}.
The latter is a query-based model.
\Cref{fig:cocorem_vitdet_mask2former_masks} shows some predicted masks by both of these models.
Through a cursory glance, we can observe that Mask2Former predicts visually sharper masks whereas masks of ViTDet have coarse boundaries.
This human judgment aligns better with COCO-ReM AP, which correctly rates Mask2Former higher.
On the other hand, COCO-2017 AP rates ViTDet higher, yielding a counter-intuitive modeling observation.

Guided by these trends, we reaffirm our claim that benchmarking newer models using the COCO-2017 may be misleading from a research perspective,
and the community should adopt our COCO-ReM annotations moving forward.
We include numerical results of \Cref{fig:cocorem_evaluations} in the appendix.


\subsection{Understanding the Difference in AP}
\label{subsec:cocorem_ap_diff}

Recall \Cref{subsec:cocorem_evaluation} (Result 1) -- all models score higher average precision (AP) on COCO-ReM, as compared to COCO-2017.
\emph{What explains the difference in AP?}
To understand the reasons underlying this difference,
we observe the components of COCO AP for a few high-performing models from \Cref{subsec:cocorem_evaluation}.
For this analysis, we consider two Cascade ViTDet models~\citep{li2022vitdet} using ViT-L/H backbones~\citep{dosovitskiy2021vit},
and two Mask2Former models~\citep{cheng2022mask2former} using Swin-B/L backbones~\citep{liu2021swin}.

\begin{figure}[t]
    \centering
    \includegraphics[width=0.99\linewidth]{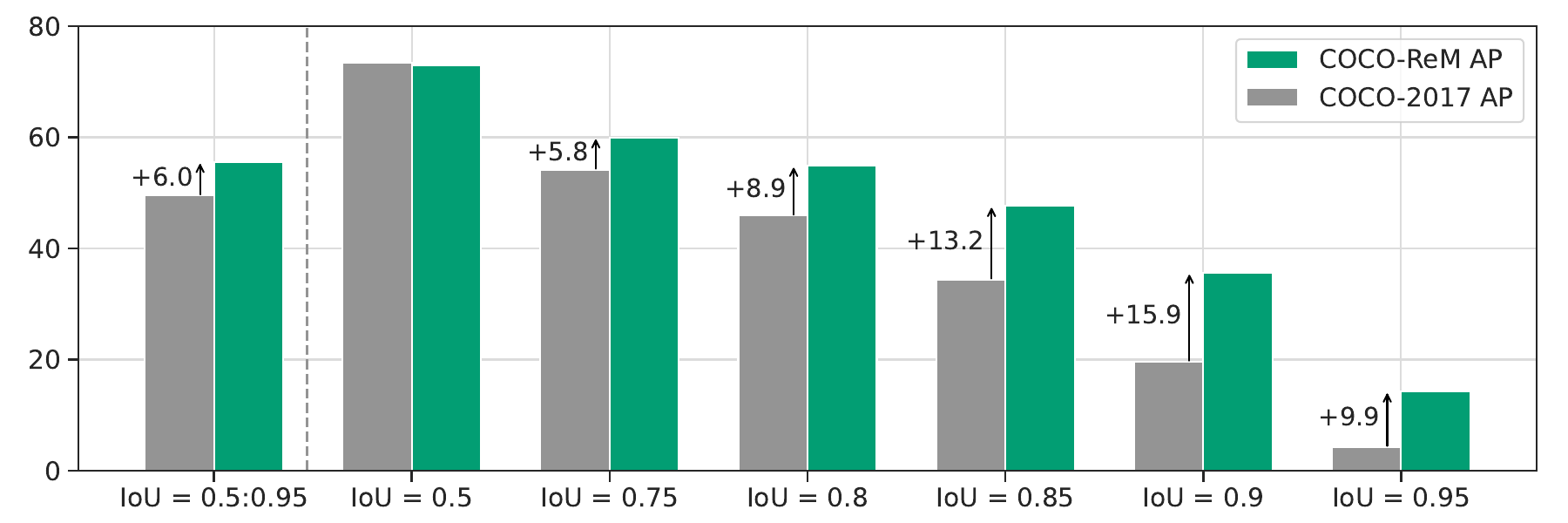}  
    \vspace{-10pt}
    \caption{
        \textbf{AP per IoU threshold.}
        While models score similarly at IoU = 0.5 on COCO-ReM and COCO-2017,
        they score much higher on COCO-ReM at higher IoU thresholds (0.75 and beyond),
        contributing to the overall difference in AP.
        Due to the improved mask quality of COCO-ReM, the AP at high IoU thresholds becomes more sensitive and enables better assessment of model capabilities.
    }
    \label{fig:cocorem_ap_diff_per_iou}
    \vspace{-5pt}
\end{figure}

\begin{figure}[t]
    \centering
    \includegraphics[width=0.99\linewidth]{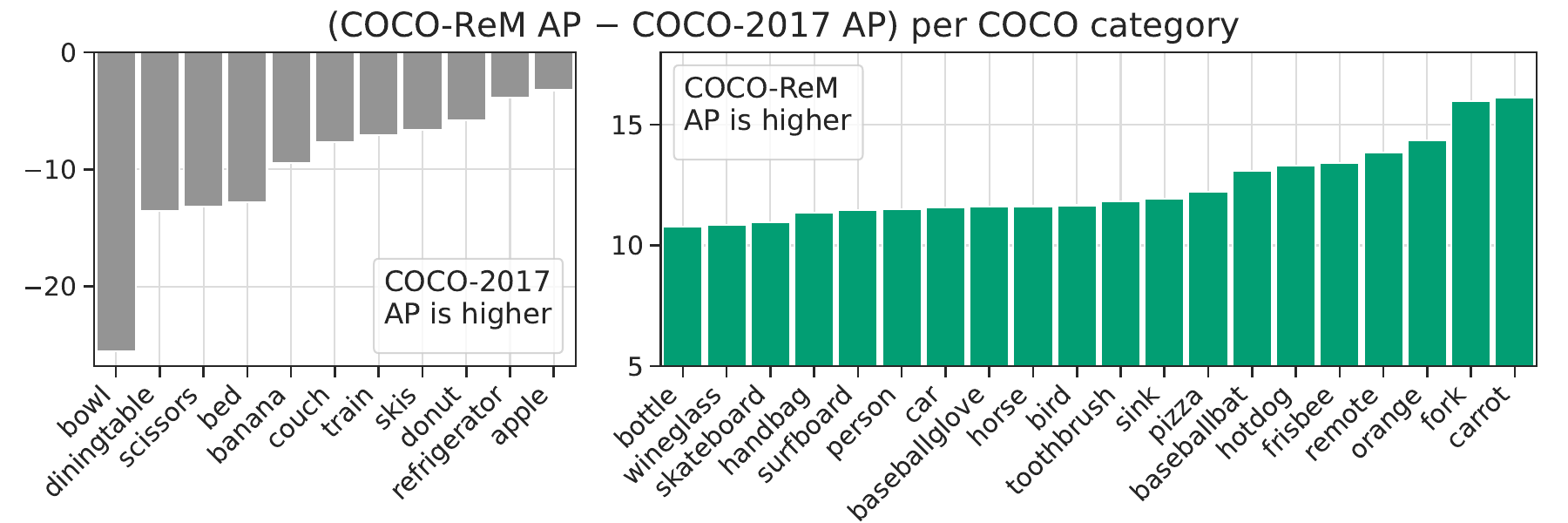}
    \vspace{-10pt}
    \caption{
        \textbf{AP per category.}
        Models score higher AP on COCO-ReM for 69 out of 80 categories.
        The masks of the remaining 11 categories (left) in COCO-2017 do not handle occlusions consistently (\emph{dining table}, \emph{bed}),
        lack holes (\emph{scissors}, \emph{donut}), or have grouped instances (\emph{banana}, \emph{apple}).
        Models learn these biases from COCO-2017 training masks and score lower COCO-ReM AP.
    }
    \label{fig:cocorem_ap_diff_per_cat}
\end{figure}

\paragraph{AP per IoU threshold.}
The calculation of COCO AP involves an average of per-category AP scores of 80 COCO categories,
each computed at ten different IoU thresholds, $t \in \{0.5, 0.55, \dots, 0.9, 0.95\}$.
In \Cref{fig:cocorem_ap_diff_per_iou}, we show components of AP at different IoU thresholds, averaged for the four models we use.
On average, models score similarly at IoU = 0.5 for COCO-ReM and COCO-2017.
However, COCO-ReM AP is much larger for IoU = 0.75 and beyond.
Hence, higher scores on COCO-ReM can be attributed to AP at higher IoU thresholds.

Intuitively, AP at IoU = $t$ rewards model predictions that overlap with a ground-truth mask of the same category with IoU $\geq t$.
AP at IoU = 0.5 requires a very \emph{coarse} overlap between the predicted and ground-truth masks,
whereas AP at high IoU thresholds have stricter overlap requirements.
Since IoU is symmetric, imprecise boundaries of ground-truth masks (COCO-2017) can wrongfully penalize valid predictions and lead to lower AP scores.
COCO-ReM fixes this issue, as evident with higher AP at high IoU thresholds.
This analysis hints that \emph{precise} masks of COCO-ReM improve the sensitivity of AP
and enable fine-grained assessment of high-performing object detectors.

\paragraph{AP per category.}
Next, we calculate AP for every COCO category by averaging across all IoU thresholds.
\Cref{fig:cocorem_ap_diff_per_cat} shows the difference in COCO-ReM AP and COCO-2017 AP per category, averaged across the same four models.
Models score higher on COCO-ReM for 69 out of 80 categories,
which is naturally reflected in the higher overall COCO-ReM AP.
We observe large improvements in categories having a generally consistent appearance (\emph{carrot} and \emph{orange}),
as well as fine-grained details (\emph{fork}).
Intuitively speaking, the larger difference in per-category AP can be attributed to the better quality of ground-truth masks in COCO-ReM.

For 11 categories, models score lower on COCO-ReM than COCO-2017.
These categories either have holes in their natural appearance
or are often heavily occluded (\emph{bowl}, \emph{dining table}, \emph{scissors}).
Masks in COCO-ReM include holes and handle occlusions more consistently than in COCO-2017.
Hence, lower per-category AP with COCO-ReM implies that the imperfections of COCO-2017 training masks impart these undesirable biases to trained models --
upon checking the model predictions, we observed that they often include occluding objects for \emph{dining table} and lack holes for \emph{scissors}.
Such predictions are not rewarded by COCO-ReM, hence incur a drop in the per-category AP.


\subsection{Ablations}
\label{subsec:cocorem_ablations}

In this section, we conduct a simple ablation study to observe the contribution of different stages in our annotation pipeline.
For the sake of simplicity, we select the same four models as chosen in \Cref{subsec:cocorem_ap_diff}.
We evaluate these models using two versions of COCO-ReM validation set:
\begin{compactenum}[\hspace{1pt}1.]
    \item \textbf{ReM-S1:}
    This version is obtained after the Stage 1 (mask boundary refinement) in our annotation pipeline.
    It comprises the same number of instances as COCO-2017, each with an improved boundary quality.
    \item \textbf{ReM-S2a:}
    This version is obtained after importing instances from the LVIS dataset -- the \emph{automatic} part of Stage 2 (exhaustive instance annotation).
    We exclude the \emph{manual} part of Stage 2 from this ablation for simplicity.
\end{compactenum}

\begin{table}[t]
    \setlength{\tabcolsep}{6pt}
    \begin{tabularx}{\linewidth}{X cccc}
    \toprule
    COCO evaluation set $\Longrightarrow$ & \cellcolor{Gray!30} 2017 & ReM-S1 & ReM-S2a & \cellcolor{ForestGreen!30} ReM (final) \\
    \midrule
    AP & \cellcolor{Gray!30} 49.6 & 57.3 & 55.6 & \cellcolor{ForestGreen!30} 55.5 \\
    Number of instances & \cellcolor{Gray!30} 36,781 & 36,781 & 41,168 & \cellcolor{ForestGreen!30} 40,689 \\
    \bottomrule
    \end{tabularx}
    \caption{
        \textbf{COCO-ReM ablations.}
        AP of four high-performing models (Cascade ViTDet-B/L and Mask2Former Swin-B/L)
        using intermediate versions from our COCO-ReM annotation pipeline.
        Trends suggest that the coarse boundary quality of COCO-2017 masks was a major source of noise in the benchmarking process.
    }
    \label{tab:cocorem_ablations}
\end{table}

\noindent Results are shown in \Cref{tab:cocorem_ablations}.
The large difference between COCO-ReM AP and COCO-2017 AP appears immediately after Stage 1 --
this finding indicates that the coarse mask boundaries are a major cause
of noise that hurts the reliability of COCO-2017 AP.
Subsequent stages incur a negligible drop in AP -- we attribute this to the increased size of the validation set,
as COCO AP is known to favor smaller evaluation sets~\citep{gupta2019lvis}.


\subsection{Training with COCO-ReM}
\label{subsec:cocorem_training}

Our evaluation study shows that COCO-ReM can serve as a useful benchmark to draw accurate modeling conclusions.
In this section, we train strong baseline models using COCO-ReM.

\paragraph{Implementation details.}
We train ViTDet models~\citep{li2022vitdet} using their official open-source implementation~\citep{wu2019detectron2}, leaving all hyperparameters unchanged.
We train two models -- Mask R-CNN ViTDet-B (111M parameters) and Cascade ViTDet-B (141M parameters).
\citet{li2022vitdet} trained this model for 100 COCO epochs across 64$\times$ A100 GPUs (batch size of 64).
To manage these costs, we implement gradient accumulation for every 8 iterations to emulate a batch size of 64 using our 8$\times$ A40 GPUs,
resulting in an 11 day-long training schedule.
We could only afford to train the \emph{base} (ViT-B) models due to limited resources.

\begin{figure}[t]
    \centering
    \includegraphics[width=\linewidth]{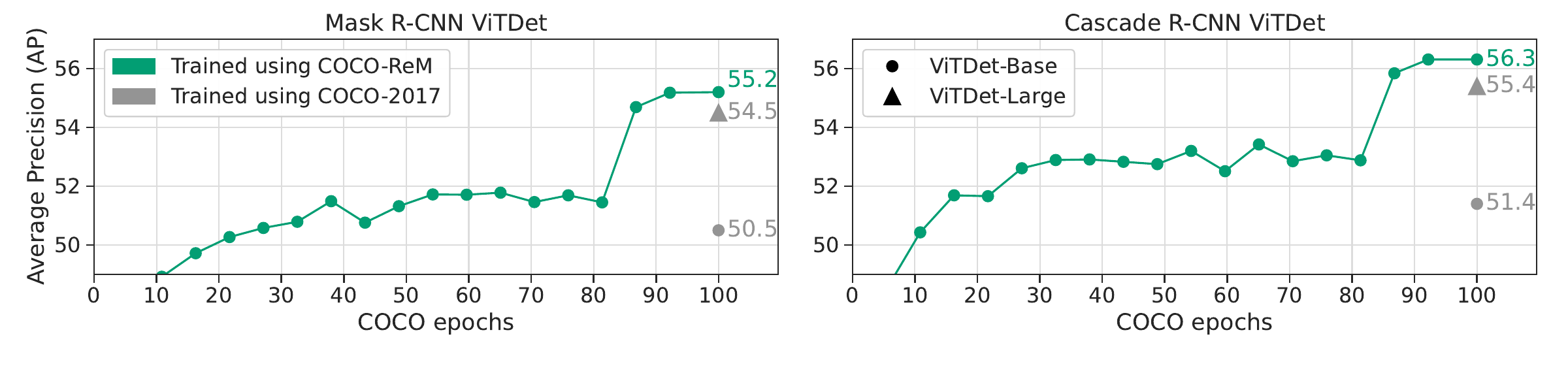}
    \vspace{-20pt}
    \caption{
        \textbf{Training with COCO-ReM.}
        A ViTDet model trained using COCO-ReM training set
        converges faster and performs better than its counterpart trained using COCO-2017,
        and closely matches a 3$\times$ larger ViTDet-L trained using COCO-2017.
    }
    \vspace{-10pt}
    \label{fig:cocorem_training}
\end{figure}

\paragraph{Results.}
\cref{fig:cocorem_training} shows that our ViTDet-B models trained using COCO-ReM significantly outperform the same ViTDet-B models trained using COCO-2017.
Higher quality of training data enables faster convergence.
Moreover, our ViTDet-B models are competitive with 3$\times$ larger ViTDet-L models trained using COCO-2017, indicating better parameter efficiency.
While numerous prior works improve object detectors through modeling innovations,
our results highlight the critical role of data quality in improving object detectors.

\section{Related Work}
\label{sec:cocorem_related}

\paragraph{Object detection and image segmentation.}
Visual perception is fundamental for any vision system to understand the world around it.
At the core of a visual perception system is the task of image segmentation that aims at recognizing the entities (objects) in a given scene and densely labeling every pixel constituting the corresponding object.
Image segmentation can broadly be divided into three sub-tasks:
semantic, instance, and panoptic segmentation.
On the one hand, semantic segmentation~\citep{fcn, deeplabv1, deeplabv2} targets obtaining a single amorphous binary mask covering all pixels for the corresponding category.
On the other hand, instance segmentation~\citep{mask-rcnn, cai2018cascade} involves detecting pixels for only foreground ``thing" regions with clear boundaries and differentiating among separate instances for a category. Panoptic segmentation~\citep{max-deeplab, panoptic-deeplab, sem-fpn} is a combination of the former two tasks targeted at predicting distinct binary masks for ``thing" regions and a single amorphous mask for the ``stuff" regions.

In this work, we particularly look at the task of instance segmentation
and contribute high-quality mask annotations to benchmark further progress.

\paragraph{Image segmentation benchmarks.}
In the last decade, considerable advancements have been observed in the classical vision task of image segmentation.
A part of this progress is owed to the introduction of numerous datasets for benchmarking image segmentation models.
In one of the earliest efforts, \citet{pascal-voc-2012} introduced the PASCAL-VOC-2012 challenge to recognize 20 classes in about 10K images.
More challenging datasets were introduced in the following years, with ADE20K~\citep{ade20k} covering 150 classes.
\citet{lin2014coco} introduced the concept of ``stuff" and ``thing" classes in their COCO dataset with a total of 171 classes: 80 ``thing" and 91 ``stuff".
The Cityscapes~\citep{cityscapes}, Mapillary-Vistas~\citep{mapillary}, and KITTI~\citep{kitti} datasets introduced image segmentation datasets focused on autonomous driving applications.

\citet{gupta2019lvis} introduced the LVIS dataset with exhaustive annotations covering 1203 categories for COCO images.
OpenImages~\citep{benenson2019lopenimages} introduced 2.8M mask annotations covering 350 classes.
More recently, the SA-1B dataset~\citep{kirillov2023sam} introduced over 1.1B fine-grained, \emph{class-agnostic} mask annotations for over 11M images.
Despite various datasets, COCO has remained the de-facto benchmark for object detectors on the instance segmentation task.
In this work, we took a closer look at the COCO instance annotations and developed a new benchmark with much higher mask quality.

\paragraph{Re-assesing datasets for benchmarking.}
Human-annotated datasets are always prone to errors and imperfections,
as no human is perfect.
Some prior works have shared our motivation and conducted in-depth re-assessment studies,
the closest to ours are works that re-assess the ImageNet dataset~\citep{recht2019imagenet,beyer2020imagenetreal}.
Concurrent to our work, \citet{zimmermann2023samacoco} also re-assess COCO annotations,
however they opt for recollecting \emph{polygon masks} from human annotators.

\section{Conclusion}
\label{sec:conclusion}

In this work, we revisit the COCO-2017 mask annotations and find major issues involving,
but not limited to their boundary quality and exhaustiveness.
These issues raise significant concerns about the dependability of COCO-2017
to benchmark recent and future object detectors accurately.
Consequently, we introduce our \textbf{COCO-ReM} benchmark consisting of high-quality mask annotations obtained using an effective \emph{semi-automatic} annotation pipeline.

On comparing the mask AP trends for existing object detectors on COCO-ReM and COCO-2017, we notice significant changes in model rankings,
notably contradicting the conclusions drawn through COCO-2017.
Moreover, training using COCO-ReM shows that paying attention to mask quality is crucial to advancing the capabilities of object detectors.
We hope that COCO-ReM will aid the future research in object detection.

\section*{Acknowledgments}

We thank Mohamed El Banani, Agrim Gupta, and Nilesh Kulkarni for helpful discussions and feedback.
We thank Palak Jagtap for her help during manual verification of masks.
We thank all the authors of the prior works for publicly releasing the pre-trained model checkpoints,
which helped us maintain empirical comprehensiveness in our evaluation study.

\clearpage

\appendix

\begin{center}
   \LARGE \textbf{Appendix}
\end{center}

This appendix consists of three parts:
(1)
\Cref{sec:cocorem_appendix_construction} includes implementation details of the mask boundary refinement stage in the COCO-ReM annotation pipeline, as discussed in \Cref{subsec:cocorem_pipeline}
(2) \Cref{sec:cocorem_appendix_dataset} includes an in-depth analysis and qualitative examples from our COCO-ReM dataset, as discussed in \Cref{subsec:cocorem_mask_chars},
and
(3) In the end, \Cref{tab:cocorem_benchmarking_full} lists the numeric results of all the fifty object detectors we evaluated in \Cref{sec:cocorem_experiments,fig:cocorem_evaluations}.

\section{Mask Boundary Refinement: Implementation Details}
\label{sec:cocorem_appendix_construction}

The first stage in our annotation pipeline -- \emph{mask boundary refinement} -- aims to refine the coarse boundaries of COCO-2017 masks.
To achieve this, we use the Segment Anything Model (SAM) for its high-quality mask predictions.
We performed an initial pass through the COCO-2017 masks using the publicly available SAM checkpoint using the ViT-H~\citep{dosovitskiy2021vit} image encoder.

\paragraph{Prompting strategy for SAM.}
We mimicked the typical interactive segmentation procedure with SAM,
following \citet{kirillov2023sam}.
We start by providing SAM with the bounding box for every COCO mask, randomly expanding by up to 10 pixels on all sides.
After the initial bounding box prompt, we input two additional point prompts randomly sampled from the error region (determined by bitwise \texttt{XOR}) between the predicted mask and the original COCO mask for SAM, labeling them foreground or background according to the corresponding pixel in the COCO mask. Following \citet{kirillov2023sam}, we provide all prior prompts and un-thresholded mask logits from the previous prompting iteration for the model to use maximal information in making a mask prediction

Since COCO masks have imprecise boundaries, randomly sampled points may be mislabeled, leading to incorrect prompt labels for SAM.
However, we observed that SAM is sufficiently robust if initially prompted with a box.
To enhance robustness, we repeat the prompting scheme ten times for every COCO mask and take a majority vote for every pixel label to obtain the final mask.

\paragraph{Manual verification.}
To further strengthen the high mask quality guarantee, we manually review every single mask obtained from the initial SAM-assisted refinement stage.
Despite a robust automated procedure, we discovered that our prompting strategy is not a \emph{one size fits all} solution.
COCO masks do not handle occlusions consistently.
Hence, SAM likely receives foreground point prompts from other occluding objects, resulting in lower-quality masks.
We found an overwhelming number of low-quality masks for six categories --
\emph{bed}, \emph{bicycle}, \emph{bowl}, \emph{dining table}, \emph{motorcycle}, and \emph{scissors}.
These usually have holes or are commonly occluded by other objects (\eg{} silverware on dining table, people and pets on beds). To accommodate these edge cases, we regenerated masks with only box (and no point) prompts to obtain refined masks for these object categories.

\section{COCO-ReM: Additional Analysis}
\label{sec:cocorem_appendix_dataset}

\begin{figure}[t]
    \centering
    \includegraphics[width=0.78\linewidth]{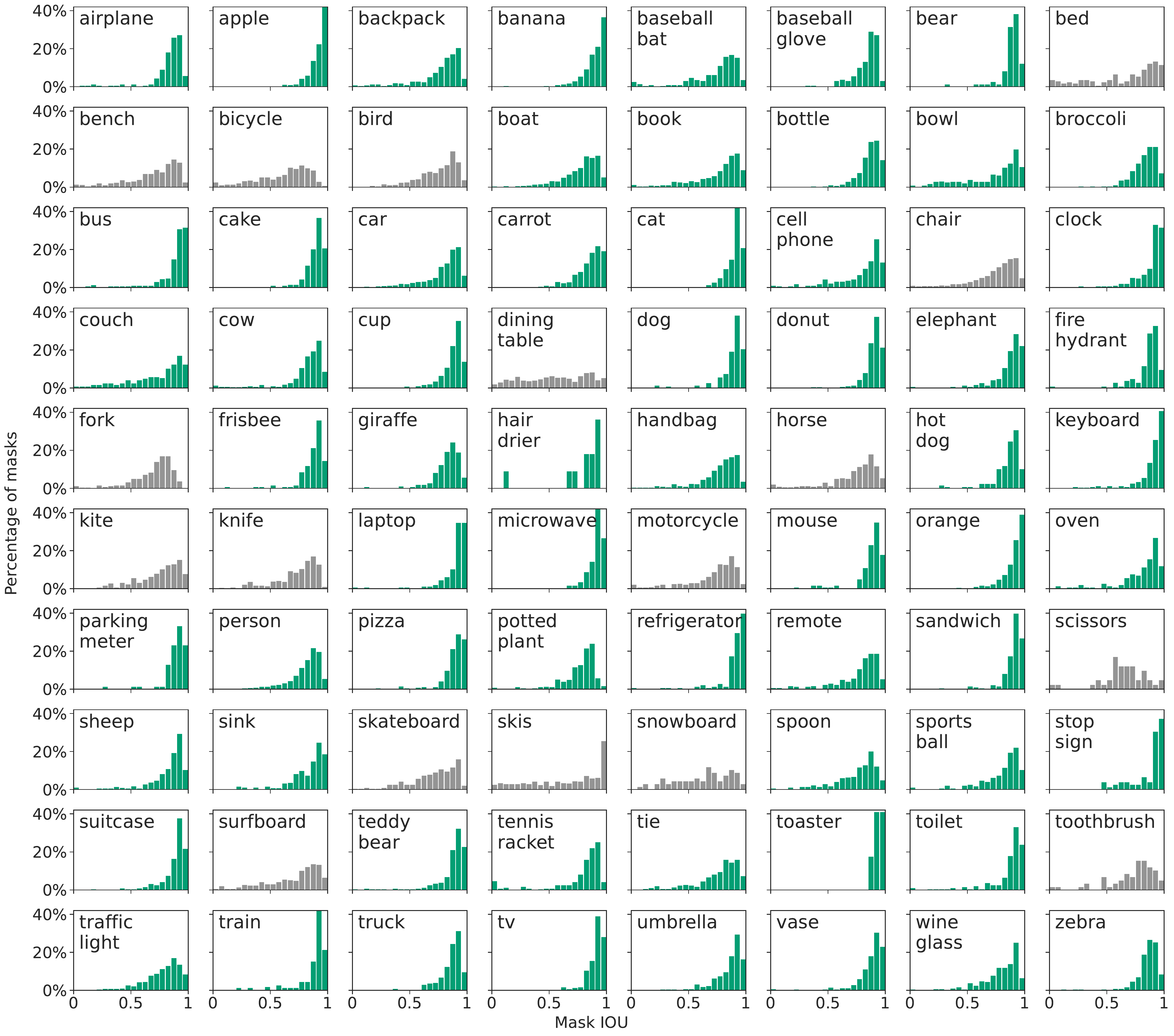}
    \vspace{-10pt}
    \caption{
        \textbf{Mask IoU per category.}
        We display the intersection-over-union (IoU) distribution between masks before and after SAM-assisted refinement and manual verification.
        Distributions with median IoU $\le 0.8$ are highlighted in gray.
    }
    \label{fig:cocorem_mask_iou_per_category}
    \vspace{-10pt}
\end{figure}


\paragraph{Per-category Mask IoU.}
In \Cref{fig:cocorem_mask_iou_per_category}, we plot the per-category mask IoU
between masks in COCO-ReM and the corresponding \emph{source} masks (from either COCO-2017 or LVIS) for the validation set.
We observe that the median of the mask IoU distribution lies close to 1 for most categories,
suggesting the change in boundary quality (enhanced) for COCO-ReM annotations. 

For the remaining classes, we notice a relatively spread-out mask IoU distribution either due to the presence of holes and occlusions in categories like \emph{scissors, dining table} or more intricate shapes for categories like \emph{knife, fork}.
Note that for masks that did not exist in the COCO-2017 or LVIS dataset,
such as those sourced from LVIS-trained models, IoU is taken to be 1.

\begin{figure}[t]
    \centering
    \includegraphics[width=0.8\linewidth]{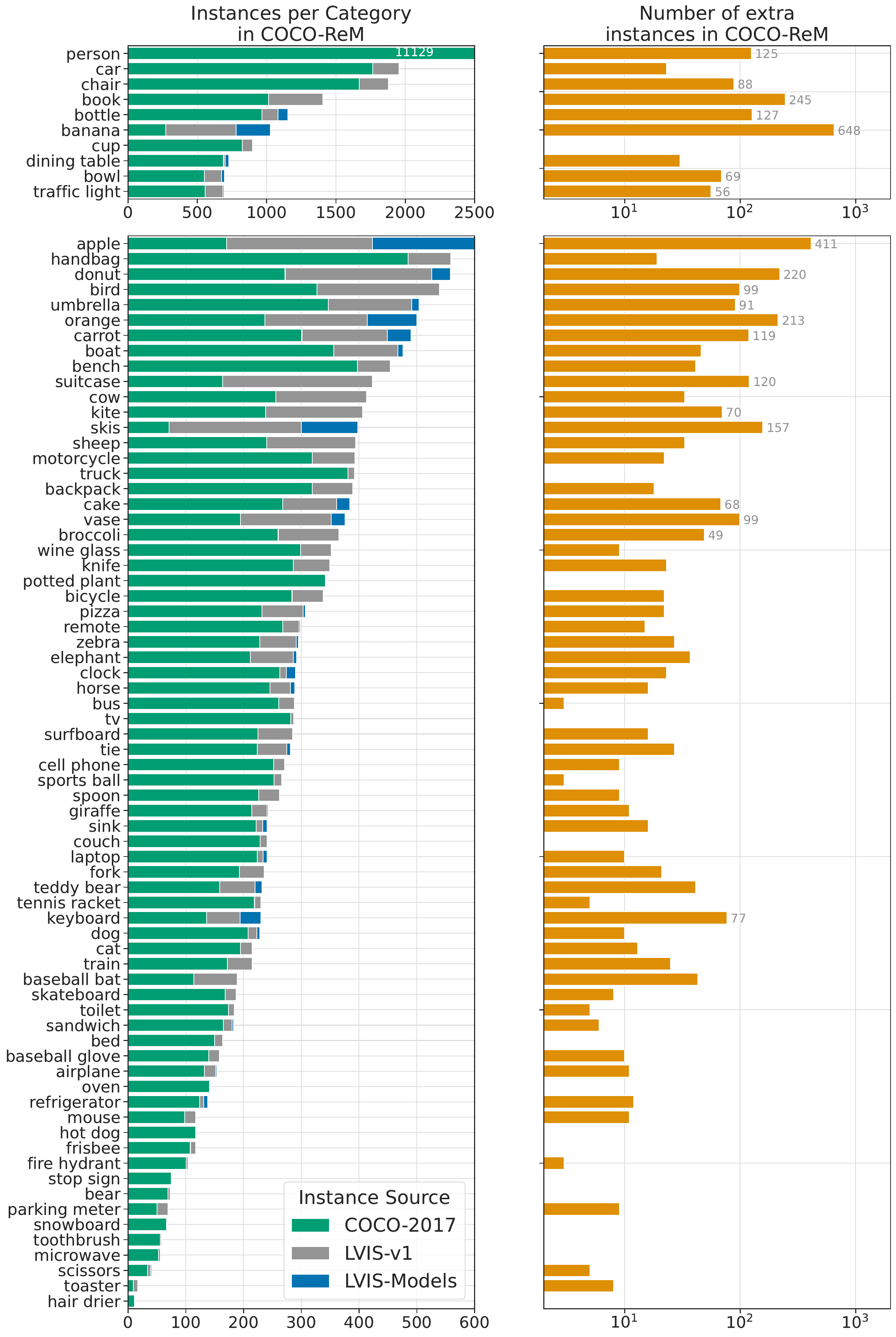}
    \caption{
        \textbf{Instance statistics in COCO-ReM.}
        \textbf{Left:} Instances per category, sourced from COCO-2017 and LVIS.
        \textbf{Right:} Number of additional masks in COCO-ReM affirm that COCO-2017 has non-exhaustive annotations.
    }
    \label{fig:cocorem_instances_per_category}
\end{figure}

\paragraph{Per-category instance statistics.}
We analyze per-category statistics about the instances' source and additional masks in our COCO-ReM evaluation benchmark as compared to COCO-2017 (validation set).
As shown in \cref{fig:cocorem_instances_per_category} \textcolor{Red}{(left)}, COCO-ReM considerably leverages the exhaustiveness quality of the LVIS~\citep{gupta2019lvis} dataset for most object categories.
Additionally, we notice that among the instances sourced from the LVIS~\citep{gupta2019lvis} dataset,
a major component is for food categories like \emph{banana, apple, orange, donut, etc.} along with categories frequently arranged collectively in an image like \emph{book, chair, etc.}, as shown in \cref{fig:cocorem_instances_per_category} \textcolor{Red}{(right)}.
We attribute these findings to the ambiguity in the annotator instructions for COCO,
resulting in non-exhaustive masks for multiple instances like a series \emph{books} in a shelf, a bunch of \emph{bananas} in a basket, or \emph{chairs} in a stadium. 

Similarly, most masks sourced from ViTDet~\citep{cai2018cascade,li2022vitdet} models trained on LVIS are also in categories like \emph{banana}, \emph{apple}, and \emph{orange}, which may have masks grouping multiple instances as they frequently occur in groups.
Therefore, unlike COCO-2017, integrating instances from LVIS and LVIS-trained models helps us maintain exhaustive instance annotations in our COCO-ReM benchmark.

\clearpage

\begin{table}
    \centering
    \scriptsize
    \renewcommand{\arraystretch}{1.0}
    \setlength{\tabcolsep}{3pt}
    \newcolumntype{Z}{>{\centering\arraybackslash}X}

    \begin{tabularx}{\linewidth}{XZZ c rrrrr c rrrrr}
        \toprule
        ~ & ~ & ~ & ~ & \multicolumn{5}{c}{COCO-2017} & ~ & \multicolumn{5}{c}{COCO-ReM} \\
        \cmidrule{5-9} \cmidrule{11-15}
        \multirow{-2}{*}{\shortstack{Detector\\Framework}} & Backbone
        & \multirow{-2}{*}{\shortstack{Pre-train\\Dataset}} & \multirow{-2}{*}{\shortstack{Training\\Epochs}}
        & AP & AP$_{50}$ & AP$_{75}$ & AP$_{90}$ & AP$_{95}$ & ~ & AP & AP$_{50}$ & AP$_{75}$ & AP$_{90}$ & AP$_{95}$ \\
        \midrule
        \rowcolor{Apricot!50} \multicolumn{15}{l}{\textbf{Source:} Detectron2~\citep{wu2019detectron2} Model Zoo (Initial and LSJ~\citep{simple-copy-paste} baselines)} \\
        \midrule
        Region &     ResNet-50 &       IN-1K &  36.0 & 37.1 & 58.5 & 39.8 & 11.2 & 1.7 & & 39.1 & 56.8 & 42.3 & 18.9 &  4.7 \\
        Cascade &     ResNet-50 &       IN-1K &  36.0 & 38.2 & 59.3 & 41.0 & 12.3 & 1.9 & & 40.6 & 57.9 & 43.8 & 20.6 &  5.5 \\
        Region &     ResNet-50 &           - & 108.0 & 39.3 & 60.8 & 42.2 & 12.7 & 2.0 & & 42.1 & 59.7 & 45.5 & 21.6 &  5.9 \\
        Region &     ResNet-50 &           - & 108.0 & 39.4 & 61.0 & 42.6 & 13.0 & 2.1 & & 42.0 & 59.4 & 45.1 & 22.1 &  6.5 \\
        \midrule
        Region &     ResNet-50 &           - & 100.0 & 40.0 & 61.7 & 43.2 & 13.5 & 2.1 & & 42.6 & 60.1 & 45.6 & 22.7 &  7.0 \\
        Region &     ResNet-50 &           - & 200.0 & 41.6 & 63.9 & 44.9 & 14.1 & 2.2 & & 44.7 & 62.4 & 48.1 & 24.2 &  7.8 \\
        Region &     ResNet-50 &           - & 400.0 & 42.3 & 64.8 & 45.6 & 15.1 & 2.5 & & 45.6 & 63.4 & 48.9 & 25.9 &  8.4 \\
        Region &    ResNet-101 &           - & 100.0 & 41.4 & 63.6 & 44.9 & 13.9 & 2.5 & & 44.4 & 61.9 & 48.1 & 24.3 &  7.5 \\
        Region &    ResNet-101 &           - & 200.0 & 42.8 & 65.5 & 46.5 & 14.8 & 2.5 & & 46.2 & 64.0 & 49.7 & 25.9 &  8.4 \\
        Region &    ResNet-101 &           - & 400.0 & 43.4 & 66.1 & 47.1 & 15.3 & 2.6 & & 47.0 & 64.6 & 50.7 & 26.8 &  8.8 \\
        \midrule
        Region &   RegNetX-4GF &       IN-1K & 100.0 & 41.3 & 63.4 & 44.7 & 13.6 & 2.5 & & 44.3 & 61.9 & 47.8 & 24.1 &  7.8 \\
        Region &   RegNetX-4GF &       IN-1K & 200.0 & 42.7 & 65.5 & 46.2 & 14.7 & 2.5 & & 46.0 & 63.8 & 49.3 & 25.9 &  8.8 \\
        Region &   RegNetX-4GF &       IN-1K & 400.0 & 43.5 & 66.4 & 47.1 & 15.3 & 2.7 & & 46.9 & 64.8 & 50.4 & 26.4 &  8.6 \\
        Region &   RegNetY-4GF &       IN-1K & 100.0 & 41.2 & 63.2 & 44.8 & 13.7 & 2.4 & & 44.2 & 61.6 & 47.7 & 23.9 &  7.5 \\
        Region &   RegNetY-4GF &       IN-1K & 200.0 & 42.7 & 65.3 & 46.1 & 15.3 & 2.4 & & 46.0 & 63.6 & 49.7 & 25.4 &  8.1 \\
        Region &   RegNetY-4GF &       IN-1K & 400.0 & 43.0 & 65.5 & 46.7 & 15.3 & 2.5 & & 46.3 & 63.8 & 50.0 & 25.9 &  8.7 \\
        \midrule
        \rowcolor{Apricot!50} \multicolumn{15}{l}{\textbf{Source:} ConvNeXt~\citep{convnext} Model Zoo} \\
        \midrule
        Region &    ConvNeXt-T &       IN-1K &  36.0 & 41.3 & 64.3 & 44.4 & 13.1 & 2.1 & & 44.1 & 63.2 & 47.8 & 21.5 &  5.3 \\
        Cascade &    ConvNeXt-T &       IN-1K &  36.0 & 43.3 & 66.2 & 46.9 & 15.5 & 2.8 & & 46.7 & 65.0 & 50.5 & 25.3 &  7.8 \\
        Cascade &    ConvNeXt-S &       IN-1K &  36.0 & 44.8 & 68.2 & 48.5 & 16.0 & 2.9 & & 48.4 & 67.2 & 52.3 & 26.1 &  8.2 \\
        Cascade &    ConvNeXt-B &       IN-1K &  36.0 & 45.4 & 68.7 & 49.2 & 16.6 & 3.2 & & 49.2 & 67.9 & 53.2 & 27.2 &  8.6 \\
        Cascade &    ConvNeXt-B &      IN-21K &  36.0 & 46.9 & 70.3 & 51.2 & 17.4 & 3.3 & & 50.8 & 69.4 & 55.3 & 28.7 &  8.9 \\
        Cascade &    ConvNeXt-L &      IN-21K &  36.0 & 47.4 & 71.2 & 51.6 & 17.6 & 3.6 & & 51.7 & 70.6 & 56.4 & 29.2 &  9.5 \\
        Cascade &   ConvNeXt-XL &      IN-21K &  36.0 & 47.7 & 71.6 & 51.9 & 18.0 & 3.4 & & 51.8 & 70.7 & 56.4 & 29.2 &  9.0 \\
        \midrule
        \rowcolor{Apricot!50} \multicolumn{15}{l}{\textbf{Source:} MViTv2~\citep{li2021mvitv2} Model Zoo} \\
        \midrule
        Region &      MViTv2-T &       IN-1K &  36.0 & 43.4 & 67.2 & 47.1 & 14.4 & 2.2 & & 47.1 & 66.4 & 51.1 & 24.8 &  7.0 \\
        Cascade &      MViTv2-T &       IN-1K &  36.0 & 44.8 & 67.9 & 48.4 & 16.0 & 3.1 & & 48.4 & 66.8 & 52.5 & 26.9 &  8.4 \\
        Cascade &      MViTv2-S &       IN-1K &  36.0 & 45.3 & 68.7 & 49.2 & 16.3 & 3.2 & & 49.5 & 68.2 & 53.6 & 28.1 &  9.5 \\
        Cascade &      MViTv2-B &       IN-1K &  36.0 & 46.5 & 70.0 & 50.6 & 17.2 & 3.4 & & 50.7 & 68.9 & 55.2 & 29.2 & 10.0 \\
        Cascade &      MViTv2-B &      IN-21K &  36.0 & 47.1 & 71.1 & 51.4 & 17.5 & 3.3 & & 51.5 & 70.0 & 55.9 & 29.9 &  9.8 \\
        Cascade &      MViTv2-H &      IN-21K &  36.0 & 48.1 & 71.8 & 52.8 & 18.5 & 3.7 & & 52.7 & 70.5 & 57.1 & 32.2 & 11.3 \\
        \midrule
        \rowcolor{Apricot!50} \multicolumn{15}{l}{\textbf{Source:} Mask2Former~\citep{cheng2022mask2former} Model Zoo} \\
        \midrule
        Query &     ResNet-50 &       IN-1K &  50.0 & 43.1 & 65.5 & 46.2 & 16.3 & 3.8 & & 47.9 & 65.2 & 51.1 & 28.7 & 11.8 \\
        Query &    ResNet-101 &       IN-1K &  50.0 & 43.7 & 66.1 & 47.0 & 16.7 & 3.9 & & 48.5 & 65.7 & 51.8 & 29.4 & 12.4 \\
        Query &        Swin-T &       IN-1K &  50.0 & 44.6 & 67.3 & 47.8 & 17.0 & 3.8 & & 50.1 & 67.4 & 53.6 & 30.9 & 12.8 \\
        Query &        Swin-S &       IN-1K &  50.0 & 46.0 & 69.2 & 49.8 & 17.5 & 3.9 & & 51.7 & 69.2 & 55.2 & 32.6 & 13.0 \\
        Query &        Swin-B &       IN-1K &  50.0 & 46.4 & 69.7 & 50.0 & 18.1 & 4.3 & & 52.2 & 69.7 & 55.8 & 32.9 & 14.2 \\
        Query &        Swin-B &      IN-21K &  50.0 & 47.6 & 71.4 & 51.5 & 18.5 & 4.4 & & 53.7 & 71.5 & 57.6 & 33.9 & 14.1 \\
        Query &        Swin-L &      IN-21K & 100.0 & 49.7 & 73.8 & 54.1 & 20.1 & 4.7 & & 56.4 & 73.8 & 60.5 & 36.6 & 16.1 \\
        \midrule
        \rowcolor{Apricot!50} \multicolumn{15}{l}{\textbf{Source:} ViTDet~\citep{li2022vitdet} Model Zoo} \\
        \midrule
        Cascade &       Swin-B &      IN-21K &  50.0 & 46.2 & 70.1 & 50.3 & 16.8 & 3.3 & & 50.5 & 69.3 & 54.8 & 28.5 &  9.0 \\
        Cascade &       Swin-L &      IN-21K &  50.0 & 47.2 & 71.2 & 51.6 & 17.5 & 3.2 & & 51.8 & 70.4 & 56.3 & 29.7 &  9.7 \\
        Cascade &     MViTv2-B &      IN-21K & 100.0 & 48.1 & 71.9 & 52.8 & 18.0 & 3.2 & & 52.7 & 70.9 & 57.2 & 31.4 & 10.7 \\
        Cascade &     MViTv2-L &      IN-21K &  50.0 & 48.3 & 72.1 & 53.0 & 18.7 & 3.5 & & 53.0 & 70.9 & 57.7 & 32.1 & 10.9 \\
        Cascade &     MViTv2-H &      IN-21K &  36.0 & 48.3 & 72.0 & 53.1 & 18.5 & 3.7 & & 53.0 & 70.8 & 57.4 & 32.5 & 11.1 \\
        \midrule
        Region &         ViT-B & IN-1K (MAE) & 100.0 & 45.9 & 69.9 & 49.8 & 16.5 & 3.0 & & 50.5 & 68.6 & 54.5 & 29.4 &  9.5 \\
        Region &         ViT-L & IN-1K (MAE) & 100.0 & 49.2 & 73.5 & 53.9 & 18.7 & 3.7 & & 54.5 & 72.6 & 59.2 & 33.5 & 12.0 \\
        Region &         ViT-H & IN-1K (MAE) &  75.0 & 50.2 & 74.5 & 55.0 & 19.3 & 3.8 & & 55.9 & 74.0 & 60.5 & 35.1 & 12.7 \\
        Cascade &        ViT-B & IN-1K (MAE) & 100.0 & 46.7 & 69.8 & 50.9 & 17.9 & 3.7 & & 51.4 & 68.7 & 55.8 & 31.0 & 11.0 \\
        Cascade &        ViT-L & IN-1K (MAE) & 100.0 & 50.0 & 73.7 & 55.0 & 19.7 & 3.9 & & 55.4 & 72.6 & 60.1 & 35.1 & 13.0 \\
        Cascade &        ViT-H & IN-1K (MAE) &  75.0 & 51.0 & 74.6 & 56.2 & 20.3 & 4.1 & & 56.7 & 73.9 & 61.7 & 36.8 & 13.7 \\
        \midrule
        \rowcolor{Apricot!50} \multicolumn{15}{l}{\textbf{Source:} OneFormer~\citep{jain2022oneformer} Model Zoo} \\
        \midrule
        Query &        Swin-L &      IN-21k & 100.0 & 48.9 & 73.5 & 53.0 & 18.6 & 4.1 & & 55.9 & 73.8 & 60.1 & 35.6 & 15.1 \\
        Query &       DiNAT-L &      IN-21k & 100.0 & 49.2 & 73.8 & 53.6 & 18.6 & 4.2 & & 56.5 & 74.3 & 60.7 & 36.8 & 15.8 \\
        Query & InternImage-H &      IN-21k & 100.0 & 52.0 & 76.8 & 57.2 & 20.8 & 4.6 & & 59.7 & 77.0 & 64.1 & 40.1 & 17.5 \\
        \bottomrule
    \end{tabularx}
    \caption{
        \textbf{Performance comparison of fifty object detectors on COCO-ReM and COCO-2017.}
        We present the numeric results of all models (trained using COCO-2017) from our evaluation study.
    }
    \label{tab:cocorem_benchmarking_full}
    \end{table}

\clearpage

\section{Datasheet for COCO-ReM dataset}
\label{sec:cocorem_datasheet}

\citet{gebru2018datasheets} introduced datasheets for datasets so that
creators can carefully reflect on the dataset curation process and
consumers can have the necessary information to make the best use of the dataset.
It includes a series of questions and answers that relay important information about
the motivation behind the dataset, its composition, collection process, maintenance, etc.
This section contains the datasheet for COCO-ReM.

\setlength{\plitemsep}{5pt}


\subsection*{Motivation}

\begin{compactenum}[\hspace{0pt}1.]
\setcounter{enumi}{0}

\item
\dsquestion{For what purpose was the dataset created? Was there a specific task in mind?
Was there a specific gap that needed to be filled? Please provide a description.} \label{Q1}
\dsanswer{This dataset was created to overcome the shortcomings of COCO-2017 (\Cref{sec:cocorem_revisit}),
and offer a set of high-quality annotations to the research community for accurate and reliable benchmarking.}

\item
\dsquestion{Who created the dataset (\eg{} which team, research group) and on behalf of which entity
(\eg{} company, institution, organization)?} \label{Q2}
\dsanswer{The annotations were created by the authors.
Images of this dataset are sourced from COCO (\url{https://cocodataset.org}).}

\item
\dsquestion{Who funded the creation of the dataset? If there is an associated grant,
please provide the name of the grantor and the grant name and number.}
\label{Q3}
\dsanswer{Creation of this dataset was not funded by any grant or person other than the authors.}

\item \dsquestion{Any other comments?}
\label{Q4}
\dsanswer{No.}

\end{compactenum}


\subsection*{Composition}

\begin{compactenum}[\hspace{0pt}1.]
\setcounter{enumi}{4}

\item
\dsquestion{What do the instances that comprise the dataset represent
(\eg{} documents, photos, people, countries)? Please provide a description.}
\label{Q5}
\dsanswer{Each instance in COCO-ReM denotes a labeled mask over a single image (from COCO).
This mask represents a grouping of pixels in an image that can be associated
with a common semantic category/label; COCO has 80 such categories.}

\item
\dsquestion{How many instances are there in total (of each type, if appropriate)?}
\label{Q6}
\dsanswer{The validation and train sets have 40,689 and 1,093,027 labeled instances, respectively.
For more details, refer to \Cref{tab:cocorem_instance_statistics,fig:cocorem_instances_per_category}.}

\item
\dsquestion{Does the dataset contain all possible instances or is it a sample of instances from a larger set?
If the dataset is a sample, then what is the larger set? Is the sample representative of the larger set
(\eg{} geographic coverage)? If so, please describe how this representativeness was validated/verified.
If it is not representative of the larger set, please describe why not
(\eg{} to cover a more diverse range of instances, because instances were withheld or unavailable).}
\label{Q7}
\dsanswer{
COCO-ReM offers a \emph{near}-exhaustive set of instance annotations for the 80 COCO categories
across all the COCO images.
We tried to ensure exhaustiveness as much as possible (refer \Cref{subsec:cocorem_pipeline}, Stage 2).
However, there there might be a tiny fraction of instances which we might have missed.
Note that COCO categories may be viewed as a subset of all possible visual concepts,
however this distinction is out of scope for COCO-ReM.}

\item
\dsquestion{What data does each instance consist of? ``Raw'' data (\eg{} unprocessed text or images) or features?
In either case, please provide a description.}
\label{Q8}
\dsanswer{Each instance comprises a segmentation mask covering an object in a single COCO image,
paired with a label out of 80 COCO category labels.}

\item
\dsquestion{Is there a label or target associated with each instance?
If so, please provide a description.}
\label{Q9}
\dsanswer{Each annotation includes a \texttt{category\_id} subfield that corresponds to a category in COCO-2017.
Each instance is also associated with a \texttt{source} (COCO-2017, LVIS dataset, or LVIS-trained models),
\texttt{source\_id}, \texttt{id}, \texttt{image\_id} and \texttt{iscrowd} field.}

\item
\dsquestion{Is any information missing from individual instances? If so, please provide a description,
explaining why this information is missing (\eg{} because it was unavailable).
This does not include intentionally removed information, but might include, \eg{} redacted text.}
\label{Q10}
\dsanswer{No, all the subfields for each instance annotation are filled with valid values.}

\item
\dsquestion{Are relationships between individual instances made explicit
(\eg{} users’ movie ratings, social network links)?
If so, please describe how these relationships are made explicit.}
\label{Q11}
\dsanswer{There exists some relationship between instances as reflected by same \emph{category\_id}.}

\item
\dsquestion{Are there recommended data splits (\eg{} training, development/validation, testing)?
If so, please provide a description of these splits, explaining the rationale behind them.}
\label{Q12}
\dsanswer{Yes, we have clearly defined training and validation sets (\Cref{tab:cocorem_instance_statistics}).
COCO test set is held privately -- we do not provide instances for the test set.}

\item
\dsquestion{Are there any errors, sources of noise, or redundancies in the dataset?
If so, please provide a description.}
\label{Q13}
\dsanswer{We use the Segment Anything Model (SAM) for generating the annotations in COCO-ReM.
SAM sometimes hallucinates small disconnected components, as reported by \citet{kirillov2023sam}.
This might be a source of noise in some instances.}

\item
\dsquestion{Is the dataset self-contained, or does it link to or otherwise rely on external resources
(\eg{} websites, tweets, other datasets)? If it links to or relies on external resources,
(a) are there guarantees that they will exist, and remain constant, over time;
(b) are there official archival versions of the complete dataset
(i.e., including the external resources as they existed at the time the dataset was created);
(c) are there any restrictions (\eg{} licenses, fees) associated with any of the external resources
that might apply to a dataset consumer?
Please provide descriptions of all external resources and any restrictions associated with them,
as well as links or other access points, as appropriate.}
\label{Q14}
\dsanswer{Images of the dataset are taken from \url{https://cocodataset.org}.
COCO is a well-established dataset in the community, and its distribution has been stable over the years.
This assures us that its availability would not be compromised.}

\item
\dsquestion{Does the dataset contain data that might be considered confidential
(\eg{} data that is protected by legal privilege or by doctor–patient confidentiality,
data that includes the content of individuals’ non-public communications)?
If so, please provide a description.}
\label{Q15}
\dsanswer{No, the dataset does not cover any information that may be considered confidential.
The images have been publicly available for nearly 10 years at the time of undertaking of this project.}

\item
\dsquestion{Does the dataset contain data that, if viewed directly,
might be offensive, insulting, threatening, or might otherwise cause anxiety? If so, please describe why.}
\label{Q16}
\dsanswer{Images of our dataset are sourced from COCO-2017 (we do not redistribute the images).
During the manual verification of our annotations, we discover eight images in the validation set (out of 5000)
to be displaying nudity. Such images might be considered sensitive in some context --
apart from this, there is no known offensive data to the best of our knowledge and perspective.}

\item
\dsquestion{Does the dataset relate to people? If not, you may skip the remaining questions in this section.}
\label{Q17}
\dsanswer{Yes, few instances in COCO-ReM as labeled as \emph{person} and they segment people in COCO images.}

\item
\dsquestion{Does the dataset identify any subpopulations (\eg{} by age, gender)?
If so, please describe how these subpopulations are identified and
provide a description of their respective distributions within the dataset.}
\label{Q18}
\dsanswer{There is no known explicit focus on any subpopulation in the dataset.
Any such bias, if discovered later, is unintentional.}

\item
\dsquestion{Is it possible to identify individuals (i.e., one or more natural persons), either directly or indirectly
(i.e., in combination with other data) from the dataset? If so, please describe how.}
\label{Q19}
\dsanswer{
COCO images contain faces of humans which are easily identifiable by naked eye.
However we do not redistribute COCO images with our dataset --
any implications of this data in COCO are not exacerbated by COCO-ReM.}

\item
\dsquestion{Does the dataset contain data that might be considered sensitive in any way
(\eg{} data that reveals race or ethnic origins, sexual orientations, religious beliefs,
political opinions or union memberships, or locations; financial or health data;
biometric or genetic data; forms of government identification, such as social security
numbers; criminal history)? If so, please provide a description.}
\label{Q20}
\dsanswer{COCO images are known to contain human faces (\qref{Q19}) and some NSFW content (\qref{Q16})
that might be considered sensitive in some context.
The images are hosted publicly and being used by the research community for nearly a decade at the time of writing.}

\item \dsquestion{Any other comments?} \label{Q21} \dsanswer{No.}

\end{compactenum}


\subsection*{Collection Process}

\paragraph{NOTE:} The authors did not collect additional data (images) beyond that available through COCO,
for developing COCO-ReM. COCO images are available at \url{https://cocodataset.org}.
Annotations from COCO-2017~\citep{lin2014coco} and LVIS~\citep{gupta2019lvis} were used.
Hence, questions in this section of the datasheet are not applicable for COCO-ReM.
We answer the questions  pertaining the annotation procedure in the next section.


\subsection*{Preprocessing/Cleaning/Labeling}

\begin{compactenum}[\hspace{0pt}1.]
\setcounter{enumi}{21}

\item
\dsquestion{Was any preprocessing/cleaning/labeling of the data done (\eg{} discretization or bucketing,
tokenization, part-of-speech tagging, SIFT feature extraction, removal of instances, processing of missing values)?
If so, please provide a description. If not, you may skip the remaining questions in this section.}
\label{Q22}
\dsanswer{We manually corrected some labeling errors in the validation set
as part of our annotation pipeline (refer \Cref{fig:cocorem_ann_pipeline}).}

\item
\dsquestion{Was the ``raw'' data saved in addition to the preprocessed/cleaned/labeled data
(\eg{} to support unanticipated future uses)?
If so, please provide a link or other access point to the ``raw'' data.}
\label{Q23}
\dsanswer{Raw data for COCO-ReM was essentially COCO (2017 version), available at \url{https://cocodataset.org}}

\item
\dsquestion{Is the software that was used to preprocess/clean/label the data available?
If so, please provide a link or other access point.}
\label{Q24}
\dsanswer{The code for the entire pipeline is available at \url{https://github.com/kdexd/coco-rem}}

\item \dsquestion{Any other comments?} \label{Q25} \dsanswer{No}

\end{compactenum}


\subsection*{Uses}

\begin{compactenum}[\hspace{0pt}1.]
\setcounter{enumi}{25}

\item
\dsquestion{Has the dataset been used for any tasks already? If so, please provide a description.}
\label{Q26}
\dsanswer{The COCO-2017 dataset has been used extensively by the research community to benchmark
vision models over the past years. COCO-ReM is a refined version of COCO-2017.
We used COCO-ReM to evaluate fifty existing models for their object detection and segmentation capabilities.
We also used COCO-ReM to train object detectors as strong baselines for future work.}

\item
\dsquestion{Is there a repository that links to any or all papers or systems that use the dataset?
If so, please provide a link or other access point.}
\label{Q27}
\dsanswer{We do not maintain such a repository. However, citation trackers like Google Scholar and
Semantic Scholar would list all future works that cite our dataset.}

\item
\dsquestion{What (other) tasks could the dataset be used for?}
\label{Q28}
\dsanswer{The dataset could be used for any machine learning task for which COCO-2017 was traditionally used.}

\item
\dsquestion{Is there anything about the composition of the dataset or the way it was collected and
preprocessed/cleaned/labeled that might impact future uses?
For example, is there anything that a dataset consumer might need to know to avoid uses that could
result in unfair treatment of individuals or groups (\eg{} stereotyping, quality of service issues)
or other risks or harms (\eg{} legal risks, financial harms)? If so, please provide a description.
Is there anything a dataset consumer could do to mitigate these risks or harms?}
\label{Q29}
\dsanswer{No, any such fact about the dataset is not evident to the authors.}

\item
\dsquestion{Are there tasks for which the dataset should not be used? If so, please provide a description.}
\label{Q30}
\dsanswer{The authors do not anticipate any task for which the use of the dataset might be harmful.}

\item \dsquestion{Any other comments?} \label{Q31} \dsanswer{No.}

\end{compactenum}


\subsection*{Distribution}

\begin{compactenum}[\hspace{0pt}1.]
\setcounter{enumi}{31}

\item
\dsquestion{Will the dataset be distributed to third parties outside of the entity
(\eg{} company, institution, organization) on behalf of which the dataset was created?
If so, please provide a description.}
\label{Q32}
\dsanswer{Yes, our dataset will be available on our dataset website, \url{https://cocorem.xyz}.}

\item \dsquestion{How will the dataset will be distributed (\eg{} tarball on website, API, GitHub)?
Does the dataset have a digital object identifier (DOI)?}
\label{Q33}
\dsanswer{We distribute our dataset as a ZIP file containing all the annotations (JSON files)
at our dataset website, \url{https://cocorem.xyz}.
Users must download the images separately from the COCO website, \url{https://cocodataset.org}}

\item
\dsquestion{When will the dataset be distributed?}
\label{Q34}
\dsanswer{The dataset will be publicly available from March 2024 onwards.}

\item
\dsquestion{Will the dataset be distributed under a copyright or other intellectual property (IP) license,
and/or under applicable terms of use (ToU)? If so, please describe this license and/or ToU,
and provide a link or other access point to, or otherwise reproduce,
any relevant licensing terms or ToU, as well as any fees associated with these restrictions.}
\label{Q35}
\dsanswer{Images of COCO-2017 are sourced from Flickr and hence should abide by Flickr terms of use
as outlined in \url{https://www.flickr.com/creativecommons/}.
The data annotation files we provide at our dataset website, \url{https://cocorem.xyz}
and code we provide at \url{https://github.com/kdexd/coco-rem} is open sourced by us.}

\item
\dsquestion{Have any third parties imposed IP-based or other restrictions on the data associated with the instances?
If so, please describe these restrictions, and provide a link or other access point to,
or otherwise reproduce, any relevant licensing terms, as well as any fees associated with these restrictions.}
\label{Q36}
\dsanswer{As stated in \qref{Q35}, the user should adhere to the Flickr terms of use,
while using the images from our dataset.}

\item
\dsquestion{Do any export controls or other regulatory restrictions apply to
the dataset or to individual instances? If so, please describe these
restrictions, and provide a link or other access point to, or otherwise
reproduce, any supporting documentation.}
\label{Q37}
\dsanswer{No.}

\item \dsquestion{Any other comments?} \label{Q38} \dsanswer{No.}
\end{compactenum}


\subsection*{Maintenance}

\begin{compactenum}[\hspace{0pt}1.]
\setcounter{enumi}{38}

\item
\dsquestion{Who will be supporting/hosting/maintaining the dataset?}
\label{Q39}
\dsanswer{The dataset is hosted at \url{https://cocorem.xyz}, and the site is maintained by the authors.}

\item
\dsquestion{How can the owner/curator/manager of the dataset be contacted (\eg{} email address)?}
\label{Q40}
\dsanswer{The contact details of authors are available on the dataset website, \url{https://cocorem.xyz}.}

\item
\dsquestion{Is there an erratum? If so, please provide a link or other access point.}
\label{Q41}
\dsanswer{There is no erratum for our initial release.
We will version all errata as future releases and document them on the dataset website.}

\item
\dsquestion{Will the dataset be updated (\eg{} to correct labeling errors, add new instances, delete instances)?
If so, please describe how often, by whom, and how updates will be communicated to dataset consumers
(\eg{} mailing list, GitHub)?}
\label{Q42}
\dsanswer{We do not have any plans to update the dataset as of now.
However, we may consider changes if deemed appropriate in the future.
Any changes will be released and communicated through our dataset website, \url{https://cocorem.xyz}.}

\item
\dsquestion{If the dataset relates to people, are there applicable limits on the retention
of the data associated with the instances (\eg{} were the individuals in question told that
their data would be retained for a fixed period of time and then deleted)?
If so, please describe these limits and explain how they will be enforced.}
\label{Q43}
\dsanswer{Yes, COCO images in the dataset contain human faces.
However, we do not control the distribution of these images (\qref{Q20}),
hence the limits to the retention of the images, if any, are the same as COCO.}

\item
\dsquestion{Will older versions of the dataset continue to be supported/hosted/maintained?
If so, please describe how. If not, please describe how its obsolescence will be communicated to dataset consumers.}
\label{Q44}
\dsanswer{
Yes, we will continue hosting all versions of COCO-ReM on our dataset website, \url{https://cocorem.xyz},
in the event of a release of newer versions.}

\item
\dsquestion{If others want to extend/augment/build on/contribute to the dataset,
is there a mechanism for them to do so? If so, please provide a description.
Will these contributions be validated/verified? If so, please describe how.
If not, why not? Is there a process for communicating/distributing these contributions to dataset consumers?
If so, please provide a description.}
\label{Q45}
\dsanswer{
Anyone who wishes to extend or contribute to the dataset in any way is welcome to contact the authors
through the provided contacts on the dataset website, \url{https://cocorem.xyz}.
All contributions and updates will be highlighted on the dataset website.}

\item \dsquestion{Any other comments?}
\label{Q46}
\dsanswer{No.}

\end{compactenum}

\bibliographystyle{plainnat}
\bibliography{main}

\begin{thebibliography}{39}
\providecommand{\natexlab}[1]{#1}
\providecommand{\url}[1]{\texttt{#1}}
\expandafter\ifx\csname urlstyle\endcsname\relax
  \providecommand{\doi}[1]{doi: #1}\else
  \providecommand{\doi}{doi: \begingroup \urlstyle{rm}\Url}\fi

\bibitem[Benenson et~al.(2019)Benenson, Popov, and Ferrari]{benenson2019lopenimages}
Rodrigo Benenson, Stefan Popov, and Vittorio Ferrari.
\newblock Large-scale interactive object segmentation with human annotators.
\newblock In \emph{Proceedings of IEEE Conference on Computer Vision and Pattern Recognition (CVPR)}, 2019.

\bibitem[Beyer et~al.(2020)Beyer, H{\'e}naff, Kolesnikov, Zhai, and Oord]{beyer2020imagenetreal}
Lucas Beyer, Olivier~J H{\'e}naff, Alexander Kolesnikov, Xiaohua Zhai, and A{\"a}ron van~den Oord.
\newblock Are we done with imagenet?
\newblock \emph{arXiv preprint arXiv:2006.07159}, 2020.

\bibitem[Cai and Vasconcelos(2018)]{cai2018cascade}
Zhaowei Cai and Nuno Vasconcelos.
\newblock {Cascade R-CNN}: Delving into high quality object detection.
\newblock In \emph{Proceedings of IEEE Conference on Computer Vision and Pattern Recognition (CVPR)}, 2018.

\bibitem[Carion et~al.(2020)Carion, Massa, Synnaeve, Usunier, Kirillov, and Zagoruyko]{detr}
Nicolas Carion, Francisco Massa, Gabriel Synnaeve, Nicolas Usunier, Alexander Kirillov, and Sergey Zagoruyko.
\newblock End-to-end object detection with transformers.
\newblock In \emph{Proceedings of European Conference on Computer Vision (ECCV)}, 2020.

\bibitem[Chen et~al.(2015)Chen, Papandreou, Kokkinos, Murphy, and Yuille]{deeplabv1}
Liang-Chieh Chen, George Papandreou, Iasonas Kokkinos, Kevin Murphy, and Alan~L. Yuille.
\newblock Semantic image segmentation with deep convolutional nets and fully connected crfs.
\newblock In \emph{ICLR}, 2015.

\bibitem[Chen et~al.(2017)Chen, Papandreou, Kokkinos, Murphy, and Yuille]{deeplabv2}
Liang-Chieh Chen, George Papandreou, Iasonas Kokkinos, Kevin Murphy, and Alan~L. Yuille.
\newblock Deeplab: Semantic image segmentation with deep convolutional nets, atrous convolution, and fully connected crfs.
\newblock In \emph{TPAMI}, 2017.

\bibitem[Cheng et~al.(2020)Cheng, Collins, Zhu, Liu, Huang, Adam, and Chen]{panoptic-deeplab}
Bowen Cheng, Maxwell~D Collins, Yukun Zhu, Ting Liu, Thomas~S Huang, Hartwig Adam, and Liang-Chieh Chen.
\newblock Panoptic-deeplab: A simple, strong, and fast baseline for bottom-up panoptic segmentation.
\newblock In \emph{Proceedings of IEEE Conference on Computer Vision and Pattern Recognition (CVPR)}, 2020.

\bibitem[Cheng et~al.(2022)Cheng, Misra, Schwing, Kirillov, and Girdhar]{cheng2022mask2former}
Bowen Cheng, Ishan Misra, Alexander~G. Schwing, Alexander Kirillov, and Rohit Girdhar.
\newblock Masked-attention mask transformer for universal image segmentation.
\newblock In \emph{Proceedings of IEEE Conference on Computer Vision and Pattern Recognition (CVPR)}, 2022.

\bibitem[Cordts et~al.(2016)Cordts, Omran, Ramos, Rehfeld, Enzweiler, Benenson, Franke, Roth, and Schiele]{cityscapes}
Marius Cordts, Mohamed Omran, Sebastian Ramos, Timo Rehfeld, Markus Enzweiler, Rodrigo Benenson, Uwe Franke, Stefan Roth, and Bernt Schiele.
\newblock The cityscapes dataset for semantic urban scene understanding.
\newblock In \emph{Proceedings of IEEE Conference on Computer Vision and Pattern Recognition (CVPR)}, 2016.

\bibitem[Cordts et~al.(2017)Cordts, Omran, Ramos, Rehfeld, Enzweiler, Benenson, Franke, Roth, and Schiele]{ade20k}
Marius Cordts, Mohamed Omran, Sebastian Ramos, Timo Rehfeld, Markus Enzweiler, Rodrigo Benenson, Uwe Franke, Stefan Roth, and Bernt Schiele.
\newblock Semantic understanding of scenes through the ade20k dataset.
\newblock In \emph{Proceedings of IEEE Conference on Computer Vision and Pattern Recognition (CVPR)}, 2017.

\bibitem[Dosovitskiy et~al.(2021)Dosovitskiy, Beyer, Kolesnikov, Weissenborn, Zhai, Unterthiner, Dehghani, Minderer, Heigold, Gelly, Uszkoreit, and Houlsby]{dosovitskiy2021vit}
Alexey Dosovitskiy, Lucas Beyer, Alexander Kolesnikov, Dirk Weissenborn, Xiaohua Zhai, Thomas Unterthiner, Mostafa Dehghani, Matthias Minderer, Georg Heigold, Sylvain Gelly, Jakob Uszkoreit, and Neil Houlsby.
\newblock {An Image is Worth 16x16 Words: Transformers for Image Recognition at Scale}.
\newblock In \emph{Proceedings of the International Conference on Learning Representations (ICLR)}, 2021.

\bibitem[Everingham et~al.(2012)Everingham, Van~Gool, Williams, Winn, and Zisserman]{pascal-voc-2012}
M.~Everingham, L.~Van~Gool, C.~K.~I. Williams, J.~Winn, and A.~Zisserman.
\newblock The {PASCAL} {V}isual {O}bject {C}lasses {C}hallenge 2012 {(VOC2012)} {R}esults.
\newblock http://www.pascal-network.org/challenges/VOC/voc2012/workshop/index.html, 2012.

\bibitem[Gebru et~al.(2018)Gebru, Morgenstern, Vecchione, Vaughan, Wallach, Daum{\'e}~III, and Crawford]{gebru2018datasheets}
Timnit Gebru, Jamie Morgenstern, Briana Vecchione, Jennifer~Wortman Vaughan, Hanna Wallach, Hal Daum{\'e}~III, and Kate Crawford.
\newblock Datasheets for datasets.
\newblock \emph{arXiv preprint arXiv:1803.09010}, 2018.

\bibitem[Geiger et~al.(2012)Geiger, Lenz, and Urtasun]{kitti}
Andreas Geiger, Philip Lenz, and Raquel Urtasun.
\newblock Are we ready for autonomous driving? the kitti vision benchmark suite.
\newblock In \emph{Proceedings of IEEE Conference on Computer Vision and Pattern Recognition (CVPR)}, 2012.

\bibitem[Ghiasi et~al.(2021)Ghiasi, Cui, Srinivas, Qian, Lin, Cubuk, Le, and Zoph]{simple-copy-paste}
Golnaz Ghiasi, Yin Cui, Aravind Srinivas, Rui Qian, Tsung-Yi Lin, Ekin~D Cubuk, Quoc~V Le, and Barret Zoph.
\newblock Simple copy-paste is a strong data augmentation method for instance segmentation.
\newblock In \emph{Proceedings of IEEE Conference on Computer Vision and Pattern Recognition (CVPR)}, 2021.

\bibitem[Gupta et~al.(2019)Gupta, Dollar, and Girshick]{gupta2019lvis}
Agrim Gupta, Piotr Dollar, and Ross Girshick.
\newblock {LVIS: A dataset for large vocabulary instance segmentation}.
\newblock In \emph{Proceedings of IEEE Conference on Computer Vision and Pattern Recognition (CVPR)}, 2019.

\bibitem[Hassani and Shi(2022)]{dinat}
Ali Hassani and Humphrey Shi.
\newblock Dilated neighborhood attention transformer.
\newblock \emph{arXiv:2209.15001}, 2022.

\bibitem[He et~al.(2016)He, Zhang, Ren, and Sun]{resnet}
Kaiming He, Xiangyu Zhang, Shaoqing Ren, and Jian Sun.
\newblock Deep residual learning for image recognition.
\newblock In \emph{Proceedings of IEEE Conference on Computer Vision and Pattern Recognition (CVPR)}, 2016.

\bibitem[He et~al.(2017{\natexlab{a}})He, Gkioxari, Doll{\'a}r, and Girshick]{he2017mask}
Kaiming He, Georgia Gkioxari, Piotr Doll{\'a}r, and Ross Girshick.
\newblock {Mask R-CNN}.
\newblock In \emph{Proceedings of IEEE International Conference on Computer Vision (ICCV)}, 2017{\natexlab{a}}.

\bibitem[He et~al.(2017{\natexlab{b}})He, Gkioxari, Dollár, and Girshick]{mask-rcnn}
Kaiming He, Georgia Gkioxari, Piotr Dollár, and Ross Girshick.
\newblock Mask r-cnn.
\newblock In \emph{Proceedings of IEEE International Conference on Computer Vision (ICCV)}, 2017{\natexlab{b}}.

\bibitem[Jain et~al.(2023)Jain, Li, Chiu, Hassani, Orlov, and Shi]{jain2022oneformer}
Jitesh Jain, Jiachen Li, MangTik Chiu, Ali Hassani, Nikita Orlov, and Humphrey Shi.
\newblock {OneFormer: One Transformer to Rule Universal Image Segmentation}.
\newblock In \emph{Proceedings of IEEE Conference on Computer Vision and Pattern Recognition (CVPR)}, 2023.

\bibitem[Kirillov et~al.(2019)Kirillov, Girshick, He, and Dollár]{sem-fpn}
Alexander Kirillov, Ross Girshick, Kaiming He, and Piotr Dollár.
\newblock Panoptic feature pyramid networks.
\newblock In \emph{Proceedings of IEEE Conference on Computer Vision and Pattern Recognition (CVPR)}, 2019.

\bibitem[Kirillov et~al.(2023)Kirillov, Mintun, Ravi, Mao, Rolland, Gustafson, Xiao, Whitehead, Berg, Lo, Doll{\'a}r, and Girshick]{kirillov2023sam}
Alexander Kirillov, Eric Mintun, Nikhila Ravi, Hanzi Mao, Chloe Rolland, Laura Gustafson, Tete Xiao, Spencer Whitehead, Alexander~C. Berg, Wan-Yen Lo, Piotr Doll{\'a}r, and Ross Girshick.
\newblock {Segment Anything}.
\newblock In \emph{Proceedings of IEEE International Conference on Computer Vision (ICCV)}, 2023.

\bibitem[Krizhevsky et~al.(2012)Krizhevsky, Sutskever, and Hinton]{imagenet}
Alex Krizhevsky, Ilya Sutskever, and Geoffrey~E Hinton.
\newblock Imagenet classification with deep convolutional neural networks.
\newblock In \emph{Advances in Neural Information Processing Systems (NeurIPS)}, 2012.

\bibitem[Li et~al.(2022{\natexlab{a}})Li, Mao, Girshick, and He]{li2022vitdet}
Yanghao Li, Hanzi Mao, Ross Girshick, and Kaiming He.
\newblock Exploring plain vision transformer backbones for object detection.
\newblock In \emph{Proceedings of European Conference on Computer Vision (ECCV)}, 2022{\natexlab{a}}.

\bibitem[Li et~al.(2022{\natexlab{b}})Li, Wu, Fan, Mangalam, Xiong, Malik, and Feichtenhofer]{li2021mvitv2}
Yanghao Li, Chao-Yuan Wu, Haoqi Fan, Karttikeya Mangalam, Bo~Xiong, Jitendra Malik, and Christoph Feichtenhofer.
\newblock Mvitv2: Improved multiscale vision transformers for classification and detection.
\newblock In \emph{Proceedings of IEEE Conference on Computer Vision and Pattern Recognition (CVPR)}, 2022{\natexlab{b}}.

\bibitem[Lin et~al.(2014)Lin, Maire, Belongie, Hays, Perona, Ramanan, Doll{\'a}r, and Zitnick]{lin2014coco}
Tsung-Yi Lin, Michael Maire, Serge Belongie, James Hays, Pietro Perona, Deva Ramanan, Piotr Doll{\'a}r, and C~Lawrence Zitnick.
\newblock {Microsoft COCO: Common objects in context}.
\newblock In \emph{Proceedings of European Conference on Computer Vision (ECCV)}, 2014.

\bibitem[Liu et~al.(2021)Liu, Lin, Cao, Hu, Wei, Zhang, Lin, and Guo]{liu2021swin}
Ze~Liu, Yutong Lin, Yue Cao, Han Hu, Yixuan Wei, Zheng Zhang, Stephen Lin, and Baining Guo.
\newblock Swin transformer: Hierarchical vision transformer using shifted windows.
\newblock In \emph{Proceedings of IEEE International Conference on Computer Vision (ICCV)}, 2021.

\bibitem[Liu et~al.(2022)Liu, Mao, Wu, Feichtenhofer, Darrell, and Xie]{convnext}
Zhuang Liu, Hanzi Mao, Chao-Yuan Wu, Christoph Feichtenhofer, Trevor Darrell, and Saining Xie.
\newblock A convnet for the 2020s.
\newblock In \emph{Proceedings of IEEE Conference on Computer Vision and Pattern Recognition (CVPR)}, 2022.

\bibitem[Long et~al.(2015)Long, Shelhamer, and Darrell]{fcn}
Jonathan Long, Evan Shelhamer, and Trevor Darrell.
\newblock Fully convolutional networks for semantic segmentation.
\newblock In \emph{Proceedings of IEEE Conference on Computer Vision and Pattern Recognition (CVPR)}, 2015.

\bibitem[maintainers and contributors(2016)]{torchvision2016}
TorchVision maintainers and contributors.
\newblock {TorchVision: PyTorch's Computer Vision library}.
\newblock \url{https://github.com/pytorch/vision}, 2016.

\bibitem[Neuhold et~al.(2017)Neuhold, Ollmann, Rota~Bulo, and Kontschieder]{mapillary}
Gerhard Neuhold, Tobias Ollmann, Samuel Rota~Bulo, and Peter Kontschieder.
\newblock The mapillary vistas dataset for semantic understanding of street scenes.
\newblock In \emph{Proceedings of IEEE International Conference on Computer Vision (ICCV)}, 2017.

\bibitem[Radosavovic et~al.(2020)Radosavovic, Kosaraju, Girshick, He, and Doll{\'a}r]{Radosavovic2020}
Ilija Radosavovic, Raj~Prateek Kosaraju, Ross Girshick, Kaiming He, and Piotr Doll{\'a}r.
\newblock Designing network design spaces.
\newblock In \emph{Proceedings of IEEE Conference on Computer Vision and Pattern Recognition (CVPR)}, 2020.

\bibitem[Recht et~al.(2019)Recht, Roelofs, Schmidt, and Shankar]{recht2019imagenet}
Benjamin Recht, Rebecca Roelofs, Ludwig Schmidt, and Vaishaal Shankar.
\newblock Do imagenet classifiers generalize to imagenet?
\newblock In \emph{Proceedings of the International Conference on Machine Learning (ICML)}, 2019.

\bibitem[Vaswani et~al.(2017)Vaswani, Shazeer, Parmar, Uszkoreit, Jones, Gomez, Kaiser, and Polosukhin]{vaswani2017attention}
Ashish Vaswani, Noam Shazeer, Niki Parmar, Jakob Uszkoreit, Llion Jones, Aidan~N. Gomez, Lukasz Kaiser, and Illia Polosukhin.
\newblock Attention is all you need.
\newblock In \emph{Advances in Neural Information Processing Systems (NeurIPS)}, 2017.

\bibitem[Wang et~al.(2021)Wang, Zhu, Adam, Yuille, and Chen]{max-deeplab}
Huiyu Wang, Yukun Zhu, Hartwig Adam, Alan Yuille, and Liang-Chieh Chen.
\newblock {MaX-DeepLab}: End-to-end panoptic segmentation with mask transformers.
\newblock In \emph{Proceedings of IEEE Conference on Computer Vision and Pattern Recognition (CVPR)}, 2021.

\bibitem[Wang et~al.(2023)Wang, Dai, Chen, Huang, Li, Zhu, Hu, Lu, Lu, Li, et~al.]{wang2022internimage}
Wenhai Wang, Jifeng Dai, Zhe Chen, Zhenhang Huang, Zhiqi Li, Xizhou Zhu, Xiaowei Hu, Tong Lu, Lewei Lu, Hongsheng Li, et~al.
\newblock Internimage: Exploring large-scale vision foundation models with deformable convolutions.
\newblock In \emph{Proceedings of IEEE Conference on Computer Vision and Pattern Recognition (CVPR)}, 2023.

\bibitem[Wu et~al.(2019)Wu, Kirillov, Massa, Lo, and Girshick]{wu2019detectron2}
Yuxin Wu, Alexander Kirillov, Francisco Massa, Wan-Yen Lo, and Ross Girshick.
\newblock Detectron2.
\newblock \url{https://github.com/facebookresearch/detectron2}, 2019.

\bibitem[Zimmermann et~al.(2023)Zimmermann, Szeto, Pasquero, and Ratle]{zimmermann2023samacoco}
Eric Zimmermann, Justin Szeto, Jerome Pasquero, and Frederic Ratle.
\newblock Benchmarking a benchmark: How reliable is ms-coco?
\newblock \emph{arXiv preprint arXiv:2311.02709}, 2023.

\end{thebibliography}
\end{document}